\documentclass[1p]{elsarticle} 
\usepackage{graphicx}

\usepackage{soul}
\usepackage{changepage,threeparttable} 
\usepackage{subcaption}
\usepackage{float}
\usepackage{booktabs}
\usepackage{flushend}
\usepackage[table,xcdraw]{xcolor}
\usepackage{multirow}
\usepackage{makecell}
\usepackage{tabularx}
\usepackage{siunitx}
\usepackage[ruled,vlined,linesnumbered, noalgohanging]{algorithm2e}
\usepackage{amsmath}
\usepackage[T1]{fontenc}
\graphicspath{ {./images/} }
\newcommand{\quotes}[1]{``#1''}

\begin{document}

\begin{frontmatter}
\title{An Instance Selection Algorithm for Big Data in High imbalanced datasets based on LSH}
\author[udea1]{Germ\'an E. Melo-Acosta\corref{cor1}}
\ead{geduardo.melo@udea.edu.co }

\author[udea1,udea2]{Freddy Duitama-Mu\~noz}
\ead{john.duitama@udea.edu.co}

\author[upm]{Juli\'an D. Arias-Londo\~no}
\ead{julian.arias@upm.es }
\cortext[cor1]{Corresponding author}
\address[udea1]{Intelligent Information Systems Lab, Universidad de Antioquia, Calle 67 No. 53 - 108, 050010, Medell\'in, Colombia.}
\address[udea2]{Department of Systems Engineering, Universidad de Antioquia, Calle 67 No. 53 - 108, 050010, Medell\'in, Colombia.}
\address[upm]{GAPS, SSR Department, ETSIT-Universidad Politécnica de Madrid, Av. Complutense, 30, 28040 Madrid, Spain.}

\begin{abstract}
Training of Machine Learning (ML) models in real contexts often deals with big data sets and high-class imbalance samples where the class of interest is unrepresented (minority class). Practical solutions using classical ML models address the problem of large data sets using parallel/distributed implementations of training algorithms, approximate model-based solutions, or applying instance selection (IS) algorithms to eliminate redundant information. However, the combined problem of big and high imbalanced datasets has been less addressed. This work proposes three new methods for IS to be able to deal with large and imbalanced data sets. The proposed methods use Locality Sensitive Hashing (LSH) as a base clustering technique, and then three different sampling methods are applied on top of the clusters (or buckets) generated by LSH. The algorithms were developed in the Apache Spark framework, guaranteeing their scalability. The experiments carried out in three different datasets suggest that the proposed IS methods can improve the performance of a base ML model between $5\%$ and $19\%$ in terms of the geometric mean.

\end{abstract}


\end{frontmatter}

\section{Introduction}

The large amount of data that currently are being generated in different fields of knowledge bring about several challenges not only for data storing and management systems but also for data analysis tools. In the Big Data context, many applications use historical data for training machine learning (ML)-based systems to provide information systems with the ability to make predictions over new data. The increment in the amount of data available for training purposes has boosted the performance in many ML applications because such an increment adds more information for the learning process, reduces the epistemic uncertainty and also helps to prevent over-fitting \cite{Halevy2009}. However, having more data is problematic for many learning algorithms since most of them have, at least, an exponential computational complexity concerning the number of training samples. Moreover, some non-parametric models use a prototype-based approach, where some training samples (or all of them) are stored in order to be used for new predictions (i.e., the well-known $k$-nearest neighbors ---$k$-NN, Support Vector Machines \cite{bishop:2006:PRML} and Gaussian Processes --- GP classification and regression methods \cite{rasmussen2006gaussian}). In those models, the predictions are based on the comparison (by applying a distance, a similarity measure or a kernel function) among the new samples and a set of the stored training samples, which implies a lot of memory and processing time \cite{arnaiz2016instance}.  \\

The specialized literature shows that there are mainly three approaches to tackle this problem: parallel/distributed implementations of ML training algorithms \cite{Chang2011},  model-based solutions that use variants of the original models and provides approximate solutions (i.e., pruned or cluster-based $k$-NN, sparse or inducing variables GPs) \cite{math8020286, Snelson2005}, and instance selection (IS) methods that reduce the number of samples to be considered during training \cite{arnaiz2016instance}. The first alternative is perhaps the most attractive one, but it requires a considerable academic community effort to get a solution for every single model that is wanted to be applied in a Big data context. Moreover, sometimes the training algorithms are, by principle, sequential; thus, the parallel/distributed variants provide approximate solutions with insurmountable performance degradation. The second alternative is similar to the first one, but its purpose is to reduce the computational complexity of the training process instead of parallelizing it. In this case, variants of the original models or training algorithms are also required to use the training data in a different and less demanding way, but most of the time, they also yield to approximate solutions. The last alternative focuses on the data used for training purposes; in this case, the solutions reduce the number of samples required to achieve proper model fitting and performance without modifying the original models or processing all the data. This strategy could also be combined with the approaches based on parallel or distributed model training. Notwithstanding their advantages, the methods in the last group must satisfy two conditions: the samples selected must represent the underlying data distribution, and their computational cost needs to be light enough to make their use worthwhile.\\

In this sense, many algorithms for IS proposed in the literature are not suitable for applications in the context of large volumes of data, due to their complexity is at least $O(n\log{n})$ or greater, and also because they are based on sorting algorithms that, by design, require all samples fit in one single machine \cite{Garcia-Osorio2010}. The computational load of these algorithms comes from the fact that usually, they must search for critical or close-to-the-decision frontier instances within the whole dataset, which results in the need to perform comparisons between each pair of instances in the datasets, or its finding depends on iterative optimization algorithms \cite{JoelLusCarboneraB2017, Garcia-Pedrajas2013}.\\ 

In addition to the computational complexity issue, ML models have to deal with a skewed data distributions in many real-life situations \cite{Krawczyk2016}, such as fraudulent transactions, mammogram lesions, or credit defaults detection. In those contexts, the severity of class imbalance can vary from minor to severe and is measured by the Imbalance Ratio(IR), computed as the number of majority class examples divided by the number of minority class examples. According to \cite{Leevy:2018, Fernandez2019}, an IR higher than 100 is considered a high imbalanced dataset; consequently, in those scenarios, the IS methods must pay attention to the sampling process of the minority class since it can undermine the possibilities to train a model properly.  In fact, some state-of-the-art works have announced that IS can help reduce the IR's impact in the training process \cite{Kuncheva2019,Garcia-Pedrajas2014}; although, no experiments in high imbalanced datasets nor their effects on the performance of ML models are thoughtfully analyzed.\\

The adaption of IS methods to tackle large data volume and high-IR problems is usually treated separately. For instance, in \cite{Garcia-Osorio2010} an algorithm called Democratic Instance Selection (DIS) is proposed, which consists in the creation of several rounds of an IS execution from different subsets of the original dataset (partitioning the training set into several disjoint subsets). The well-known Decremental Reduction Optimization Procedure Version 3 (DROP3) \cite{Wilson2000} or Iterative Case Filtering (ICF) \cite{Brighton2002} IS algorithm ``votes'' for the instances to remove in each round. Then, these rounds are combined using a voting scheme that uses a voting threshold computed with an optimization process for both the training error and the size of instances in memory. Using a similar idea of splitting the original dataset to scale the IS methods, \cite{arnaiz2016instance} proposed the use of Locality Sensitive Hashing (LSH) to create partitions of the original dataset, taking advantage of that LSH has a linear cost. The Authors used LSH partitions to create two new IS algorithms. The first one is called LSH-IS-S, it leaves only one instance of each class inside each partition and the second one called LSH-IS-F, which consists on to evaluate whether there is only one instance of a class within the partition, the instance is rejected; otherwise, if two or more instances of the same class are present, one of them is randomly chosen. In the same way, \cite{JoelLusCarboneraB2017} built an algorithm called XLDIS, which splits the original dataset using a modified $k$-NN method. Then, in each subset, the most representative instances of each class are selected using the local density of the instances (average similarity between the instance and all its pairs). Although the results demonstrated that XLDIS time complexity is lower than other traditional IS approaches, the experimental setup did not use datasets with more than $20000$ samples.\\ 


In the three previous works, although the database used had some level of imbalance, their IR was not analyzed, and their main results were focused on demonstrating the near-linear computational complexity of the algorithms. Furthermore, although the authors claimed that the proposed algorithms are suitable to implement in Big data contexts, only  DIS  was developed to be implemented using a distributed processing framework \cite{Arnaiz-Gonzalez2017}.\\

On the opposite, \cite{Czarnowski2018}, and \cite{Tsai2019} proposed new IS algorithms for high IR problems (varying from 2 to 130). Both works used similarity-based clustering algorithms (such as k-means or affinity propagation) to group the instances of each class separately; then, within the clusters of the majority class, they performed the IS. The first method used a parallelizable agent-based optimization algorithm to select the instances that maximized the problem's accuracy. However, though the algorithm can be parallelizable, it must complete multiple training loops of the same classifier to optimize its accuracy. The second method applied standard DROP3, Instance-Based Learning version 3 (IB3) \cite{DavidW.Aha1911}, and Genetic Algorithms inside the clusters (which were created only for the majority class samples) to perform the IS process. Additionally, both works focused on small datasets (minor than $5000$ instances), and the complexity of the techniques was not analyzed.\\


To the best of our knowledge, only \cite{Garcia-Pedrajas2013} aimed to adapt IS to solve the two pointed out problems. The authors proposed the Oliogaric IS (OligoIS) algorithm, introducing two modifications to the DIS algorithm: using a random dataset splitting to create the partitions and changing the voting threshold favoring minority class instances. However, even though the results showed better accuracy when the IR of the problem was higher and the algorithm was tested in a large dataset (more than $5000000$ instances), no distributed processing framework was used. Moreover, the final performance of the models trained with the reduced dataset depends on the IS method used. The authors showed that the difference in performance between the algorithm using an evolutionary IS-based method and DROP3 achieves even a $20\%$ in relative terms, which could be explained because the work did not propose any adaptation of the IS methods to the splitting process. \\


Bearing this in mind, the present work proposes three new IS algorithms to work in high IR problems, which can also be used in a Big data context. One method balances the trade-off between fast calculations and a basic informed IS process using an entropy measure to guide the amount of sampling required. The other two methods adapt the popular DROP3 algorithm to deal with imbalanced class problems and perform a more informed IS. As in \cite{arnaiz2016instance}, the proposed algorithms use LSH as a base clustering method to achieve linear computational cost in large-volume datasets, and subsequently apply the IS methods in a divide-and-conquer manner.  The influence of three LSH families is studied concerning the models' performance obtained in several experiments with different datasets. IS algorithms are also analyzed in terms of computational complexity and IR. All the algorithms were implemented using the distributed computing framework Apache Spark \cite{spark}, and are publicly accessible on Gitlab\footnote{https://gitlab.com/germanEduardo/lsh-analysis/}. \\

The rest of the paper is organized as follows. First, section \ref{section:methods} presents the LSH concepts and their families, as well as a detailed explanation of the three algorithms introduced for IS based on LSH and how they are used to solve the class imbalance problem. Secondly, section \ref{section:experiments}  describes the experimental setup, the results obtained, and their analysis. Finally, Section \ref{section:conclusions} presents the main conclusions of the work and propositions of further lines.

\section{Methods}\label{section:methods}

\subsection{Instance Selection using Locality Sensitive Hashing}\label{section:lsh}

\begin{figure}[h!]
\centering
	\includegraphics[scale=0.14]{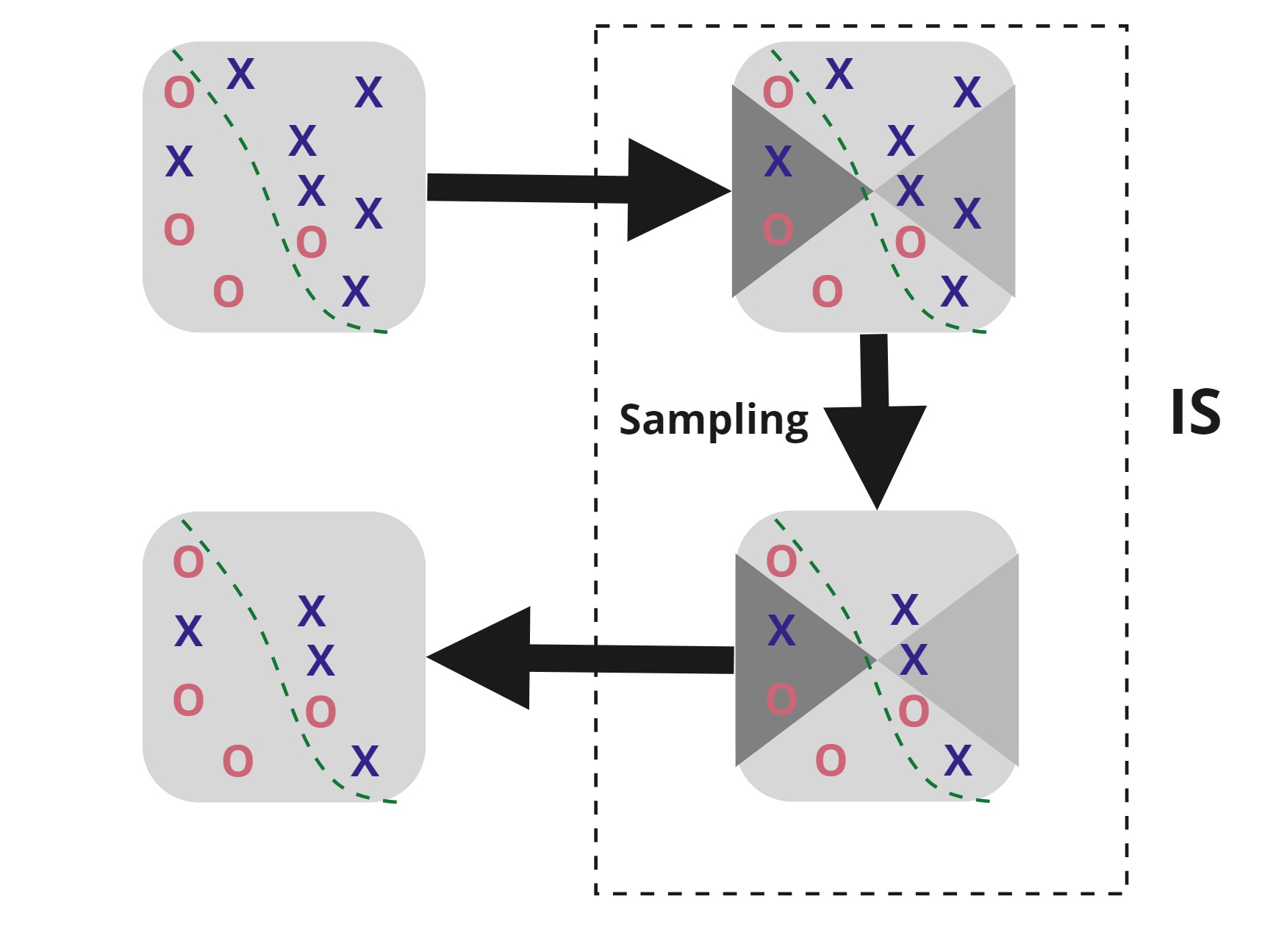}
	\caption{Methodology proposed to use LSH to perform Instance Selection}
	\label{Fig:LSH_IS_Methodology}
\end{figure}

LSH is a method for determining which items (e.g., samples on a features space) in a given set are similar and  is based on the simple idea that, if two points are close together, then after applying a \quotes{projection} operation these two points will remain close together \cite{Indyk1998}. Due to the use of hash functions instead of distances to estimate the similarity of the objects, LSH can create clusters or groups of similar objects with linear computational complexity. Figure \ref{Fig:LSH_IS_Methodology} shows a simplified diagram to demonstrate how LSH can be used to perform IS. The first step is to divide the feature space of the training data into regions using \quotes{buckets} created using LSH. The LSH buckets group the instances based on their similarity.  Next, within every bucket, a sampling algorithm is applied to select the most representative instances. The sampling techniques proposed in this work (Sections \ref{section:entropy} and \ref{section:drop3}) try to preserve the most significant instances, considering the particular challenges of high IR problems discussed before.\\ 

As mentioned, the LSH needs the definition of hash functions with particular properties. Additionally, it is essential to understand how those functions divide the feature space to raise awareness of the high-class imbalance. Ultimately, the selection of those functions determines which samples are being considered by the underline sampling method. Therefore, the following sections formalize the hash functions properties, how they are addressed in the study, and the sampling methods proposed.

\subsection{Locality-sensitive functions}\label{section:lsh-functions}

\begin{figure}[h!]
\centering
	\includegraphics[scale=0.65]{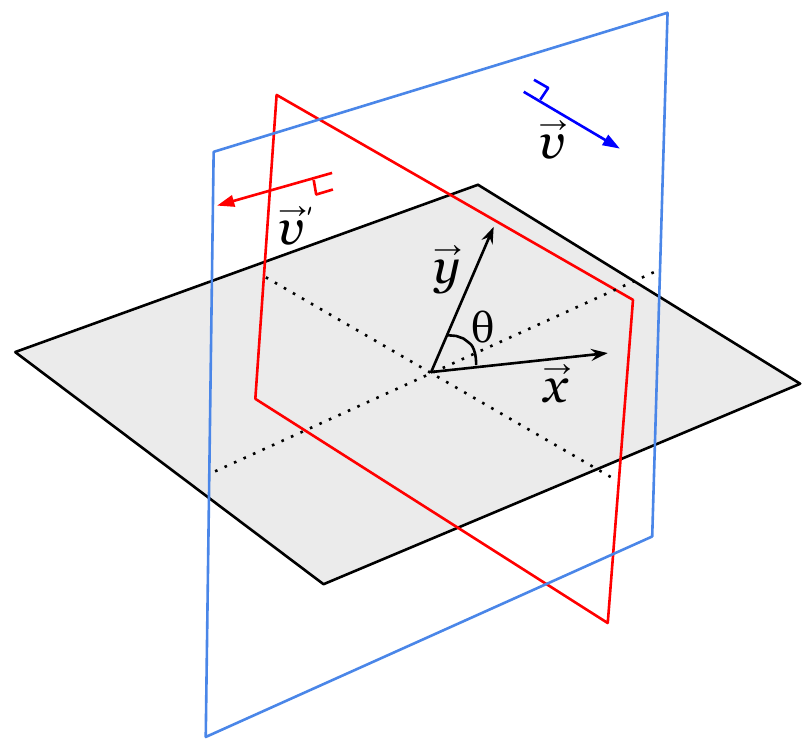}
	\caption{Two vectors $\vec{v}$ and $\vec{v'}$  and two random hyperplanes.}
	\label{Fig:Math_LSH}
\end{figure}

In the LSH approach, the central point is defining a hash function suitable for \quotes{hashing} the items under analysis. The functions to be used in LSH must take two items and provide a decision about whether these items are likely to be similar or not. These kind of functions are called locality-senstive functions. In Figure \ref{Fig:Math_LSH}, although vectors $\vec{x}$ and $\vec{y}$ may be in a space of many dimensions, they always define a plane and make an angle $\theta$ between them which is measured in this plane (gray region). Suppose we pick two additional hyperplanes through the origin. Each hyperplane intersects the plane of  $\vec{x}$  and $\vec{y}$  in a line (dotted line). To pick a random hyperplane, we actually pick the normal vector to the hyperplane, say $\vec{v}$. Note that with respect to the blue hyperplane, the vectors $\vec{x}$  and $\vec{y}$ are on different sides of this; as a consequence, the dot products $\vec{v} \cdot \vec{x}$  and $\vec{v} \cdot \vec{y}$ will have different signs. In the same way, the randomly chosen vector$\vec{v'}$ is normal to the red hyperplane. In that case, both  $\vec{v'} \cdot \vec{x}$ and $\vec{v} \cdot \vec{y}$  have the same sign. It is worth noting that all possible angles for the line that is the intersection of the random blue hyperplane and the plane of $\vec{x}$ and  $\vec{y}$ are equally likely. Thus, the blue hyperplane will choose with probability equal to $\theta/180^{\circ}$. As result, the probability that the randomly chosen vector  $\vec{v}$ places the vectors $\vec{x}$ and $\vec{y}$ on the same side of the hyperplane is $(180^{\circ} - \theta) / 180^{\circ}$.

Each hash function $f$ in our locality-sensitive family $\mathcal{F}$ will be built from a randomly chosen vector $\vec{v}$, where given two vectors $\vec{x}$  and $\vec{y}$ : $f(\vec{x})=f(\vec{y})$ if and only if the dot products  $\vec{v} \cdot \vec{x}$ and $\vec{v} \cdot \vec{y}$ have the same sign.

\begin{figure}[h!]
\centering
	\includegraphics[scale=0.5]{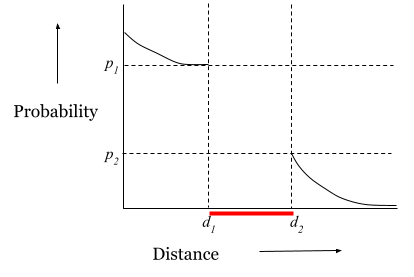}
	\caption{$(d_{1},d_{2},p_{1},p_{2})$-sensitive function}
	\label{Fig:LSH_FUNC}
\end{figure}

More formally, see Figure \ref{Fig:Math_LSH}, A locality-sensitive family $\mathcal{F}$  is $(d_{1},d_{2},p_{1},p_{2})$-sensitive function:
\begin{itemize}
    \item If $\theta(\vec{x},\vec{t})) \leq  d_{1} = \theta _{1}$, then $\mathcal{P}[f(\vec{x})=f(\vec{y})] \geq p_{1} = (180^{\circ} - \theta _{1})/ 180^{\circ}$. The function $f$ makes $\vec{x}$ and $\vec{y}$ a candidate pair to be near neighbors.
    \item If $\theta(\vec{x},\vec{t})) \geq  d_{2} = \theta _{2}$, then $\mathcal{P}[f(\vec{x})=f(\vec{y})] \leq p_{2} = (180^{\circ} - \theta _{2})/ 180^{\circ}$. The function $f$ does not make $\vec{x}$ and $\vec{y}$ a candidate pair to be near neighbors.
    \item There are no guarantees about the fraction of falses positives in the interval ($d_{1}=\theta _{1},d_{2}=\theta _{2}$) (the red line in Figure \ref{Fig:LSH_FUNC}).
\end{itemize}

A family of hash functions can be easily extended through AND and OR constructions, by taking $k$ members of $\mathcal{F}, (f_{1}; f_{2}; :::; f_{k})$ and applying the following rule:

\begin{itemize}
\item For the AND construction we say $f(x)=f(y)$ if $f_i(x) = f_i(y)$ for each $i=1,2,...,l$
\item For the OR construction we say $f(x) = f(y)$ if $f_i(x) = f_i(y)$ for one or more values $i$.
\end{itemize}

\begin{figure}[!h!b]

	\begin{subfigure}[b]{0.4\textwidth}
		\centering
		\includegraphics[scale=0.22]{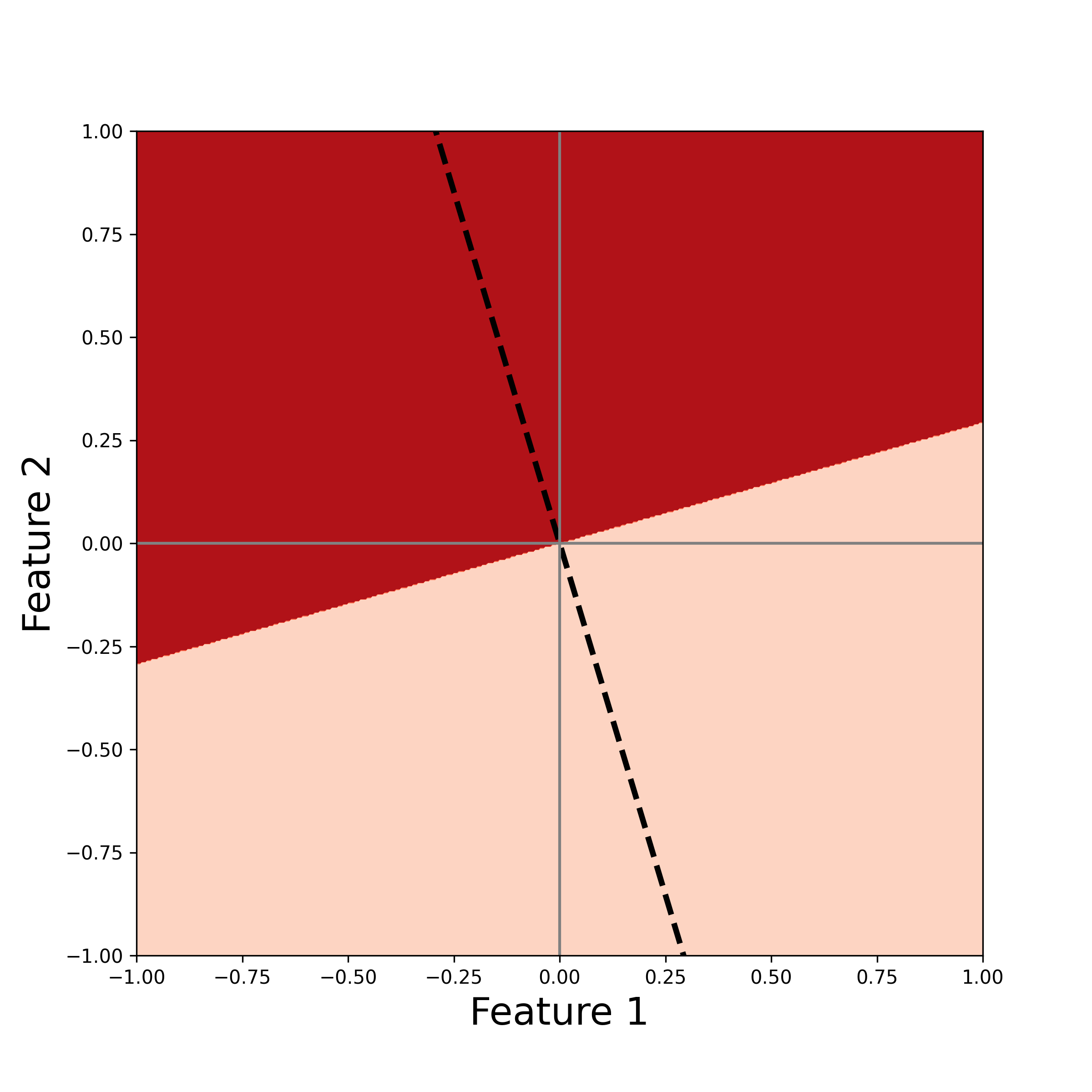}
		\caption{Regions with RHF using 1 ANDS. The Black line represents the random hyperplane generated}
		\label{Fig:HFunctiona}
	\end{subfigure}
\hfill
	\begin{subfigure}[b]{0.4\textwidth}
		\centering
		\includegraphics[scale=0.22]{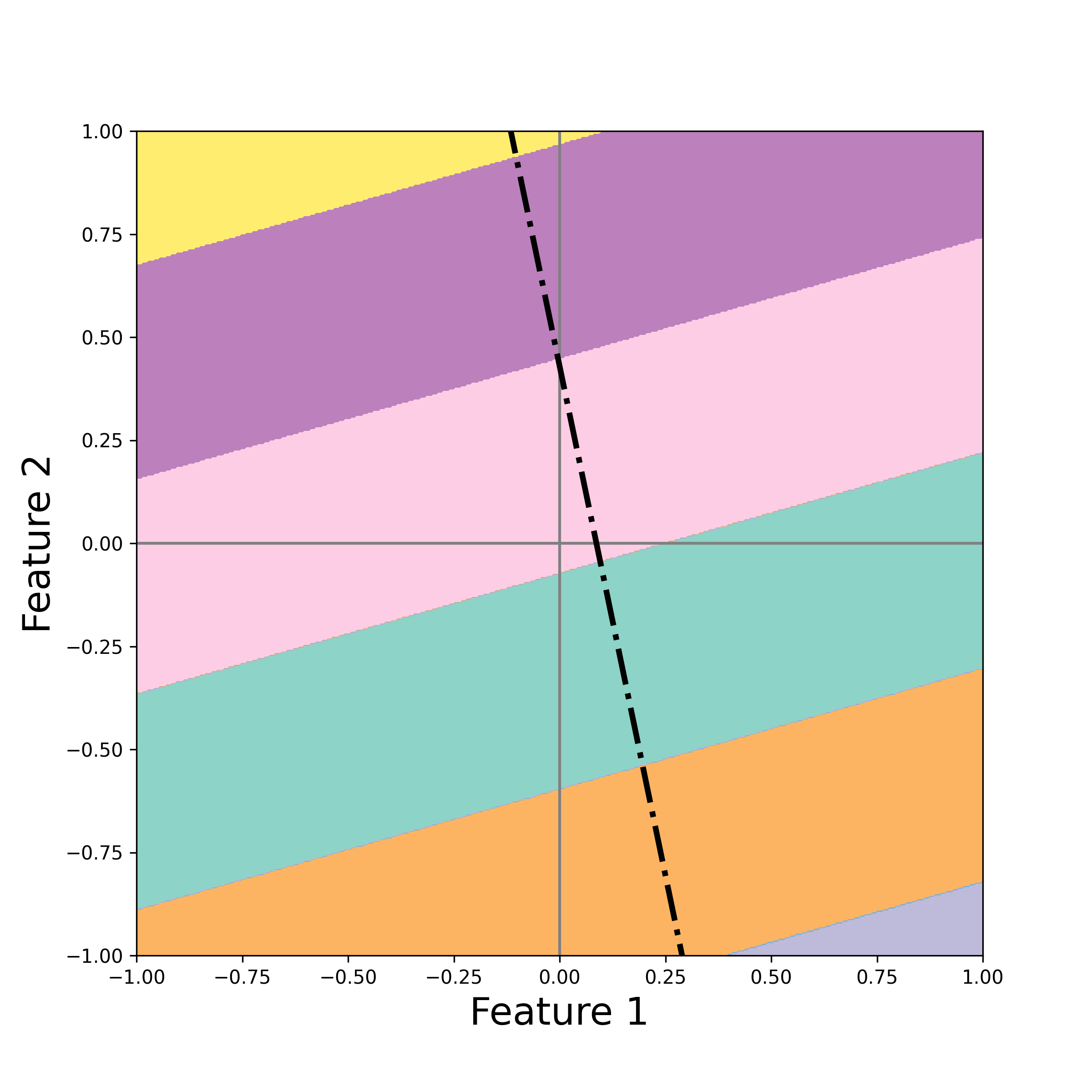}
		\caption{Regions with DPF using 1 ANDS. The Black line represents the distributions generated using $f({\bf{x}}) = \lfloor{\frac{{\bf{a}}\cdot{\bf{x}}+b}{r}} \rfloor$}
		\label{Fig:HFunctionb}
	\end{subfigure}
\caption{}
\label{Fig:Function}
\end{figure}

There are two common hash functions that can be used in this case: the random hyperplanes for the cosine distance (RHF) \cite{rajaraman2012mining} and the dot product with a random vector extracted from a p-stable distribution $f({\bf{x}}) = \lfloor{\frac{{\bf{a}}\cdot{\bf{x}}+b}{r}} \rfloor$ (DPF) \cite{datar2004locality}, where ${\bf{a}}$ is the random vector, ${\bf{x}}$ is the sample to be clustered, $r$ is the size of the bins defining every bucket, and b is drawn uniformly from $[0,1]$.  In order to split the feature space correctly, usually, both the RHF and the DPF are used according to an AND construction \cite{arnaiz2016instance}. In this sense, as long as the number of used functions $l$ increases, the value of $p_1$ increases as well, given more certainty about the similarity among the samples in a bucked, although it also has a consequence in terms of the computational load. The RHF splits the feature space into regions divided by the vectors at $90^{\circ}$ and $270^{\circ}$ from the hyperplane used (see Figure \ref{Fig:HFunctiona}), starting from the origin of the space and extending the region to infinity along the hyperplanes, while the DPF splits the feature space in bands along the random vectors, which corresponds to the normal vector of the boundary planes dividing the bands (see Figure \ref{Fig:HFunctionb}).Although both of them provide regions where it is possible to get a reasonably good clusterization, an alternative approach could be obtained by combining both families of hash functions and their properties.\\

\begin{figure}[!h!t]
\centering
	\includegraphics[scale=0.28]{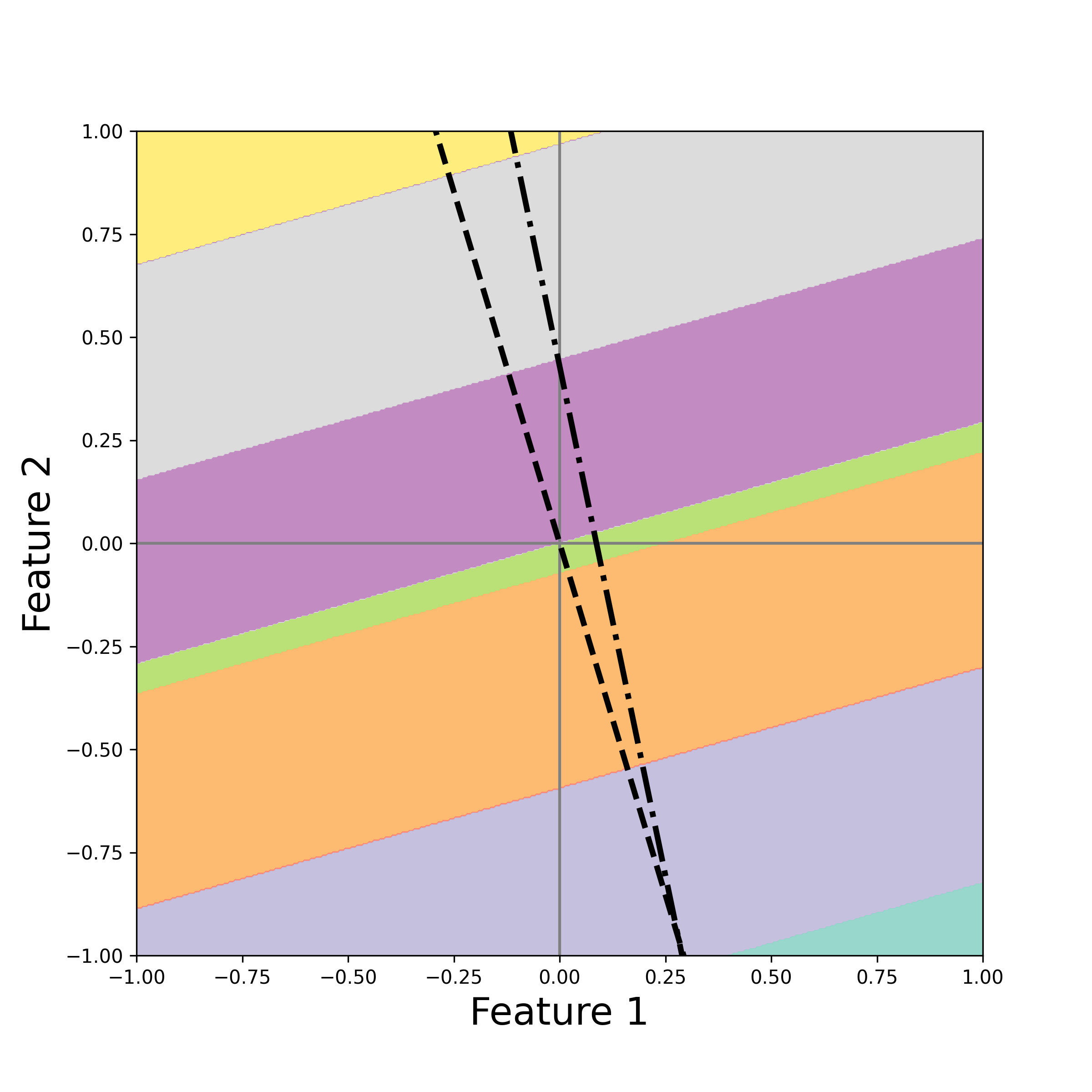}
	\caption{Regions built with the combination of RHF and DPF. The black lines represents the random hyperplane and the p-stable distribution to generated the buckets}
	\label{Fig:HFunction2}
\end{figure}

There are two common ways to combine families of hash functions: composition and concatenation \cite{ivanchykhin2017regular}. The composition method corresponds to the application in a cascade of one family of hash functions after the other. On the other hand, the concatenation is similar to the application of the AND construction, but considering functions extracted from two families $\mathcal{F}$ and $\mathcal{G}$. Taking into account that the RHF provides binary outputs whiles the DPF returns integers, it is possible to exploit the associative and distributive properties of the AND operator and to define the new hash function $h(x) = (f(x),g(x))$, as the concatenation of a family of $l$ functions $f(x)=(f_1(x),f_2(x),...,f_l(x))$, and $m$ functions $g(x) = (g_1(x),g_2(x),...,g_m(x))$. Therefore, the AND operator is firstly applied inside every family of functions and them between the two families. Figure \ref{Fig:HFunction2} shows the effect of the concatenation of RHF and DPF using the same random hyper-planes for both families ($m=l$).\\

\subsection{Entropy Sampling Method}\label{section:entropy}

The first sampling method proposed is called Entropy-based sampling. This method tries to compensate for the trade-off between fast calculations and the selection of the samples informedly using the entropy value of the samples within the region. For bi-class problems the entropy takes values in the interval $[0,1]$, so its value is interpreted as the required proportion to randomly selects samples of the majority class.  The algorithm randomly selects one unique sample when a bucket only has samples of one class (entropy=$0$). Moreover, the method does not eliminate samples belonging to the minority class due to the critical impact of dropping a minority class instance in high IR problems.  The pseudo-code of the method is presented in the Algorithm \ref{alg:lsh_entropy} including the LSH partitions creation; it uses three input parameters: the dataset, the hash function family, and the number of ANDs functions.\\

\begin{algorithm}[h!t!]
    \DontPrintSemicolon
	\SetKwData{Left}{left}\SetKwData{This}{this}\SetKwData{Up}{up}
	\SetKwFunction{Union}{Union}\SetKwFunction{FindCompress}{FindCompress}
	\SetKwInOut{Input}{input}\SetKwInOut{Output}{output}
    
    \Input{$X$ = instances dataset, $G$ = a hash function family, $A$ = number of ANDS functions}
	\Output{$S$ = selected instances $S \subset X$}
    $S \longleftarrow \emptyset$\;
    $U \longleftarrow \emptyset$\;
    \ForEach{instance $x \in X$}{ \label{alg:lsh_entropy:loop1}
    	\ForEach{hash function $m \in G$ with $A$ AND functions} {
    		$u \longleftarrow$ signature of the bucket assigned to $x$ by $m$\;
    	}
        $b \longleftarrow$ concatenation of the instance bucket signatures\;
        add bucket $b$ to $U$\;
    } \label{alg:lsh_entropy:loop2}
    \ForEach{bucket $u \in U$} {
     	calculate the entropy $H$ of $u$\;
     	
     	$s1 \longleftarrow$ all instances of the minority class\;
        
        \eIf{$H\neq 0$}{
            $s2 \longleftarrow \%$ of majority class instances in the dataset equal to $H$\;
        }{
    
            $s2 \longleftarrow$ a single one instance randomly selected\;
        }
    $s \longleftarrow s1 \cup s2$\;
        
    add $s$ to $S$\;
    }
    \Return $S$ 
 \caption{Instance selection with LSH Entropy-Based Sampling using 1 ANDS}
 \label{alg:lsh_entropy}
\end{algorithm}

\begin{figure*}[!h!t]
	\begin{subfigure}[b]{0.33\textwidth}
		\centering
		\includegraphics[scale=0.2]{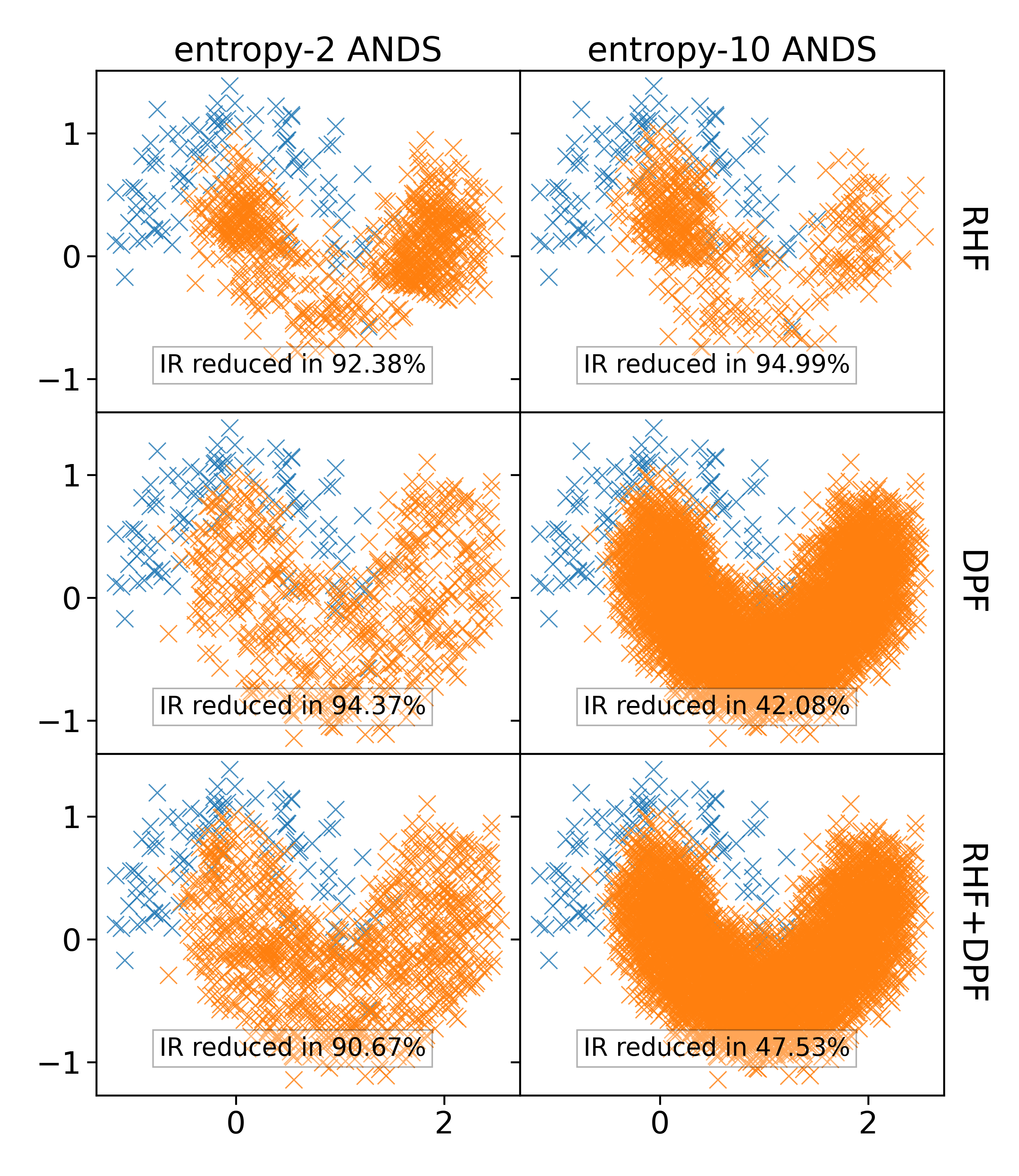}
		\caption{Effect of Entropy-based sampling in an interleaving half-circles dataset with $IR=100$}
		\label{Fig:sampling_example1a}
	\end{subfigure}
\hfill
	\begin{subfigure}[b]{0.33\textwidth}
		\centering
		\includegraphics[scale=0.2]{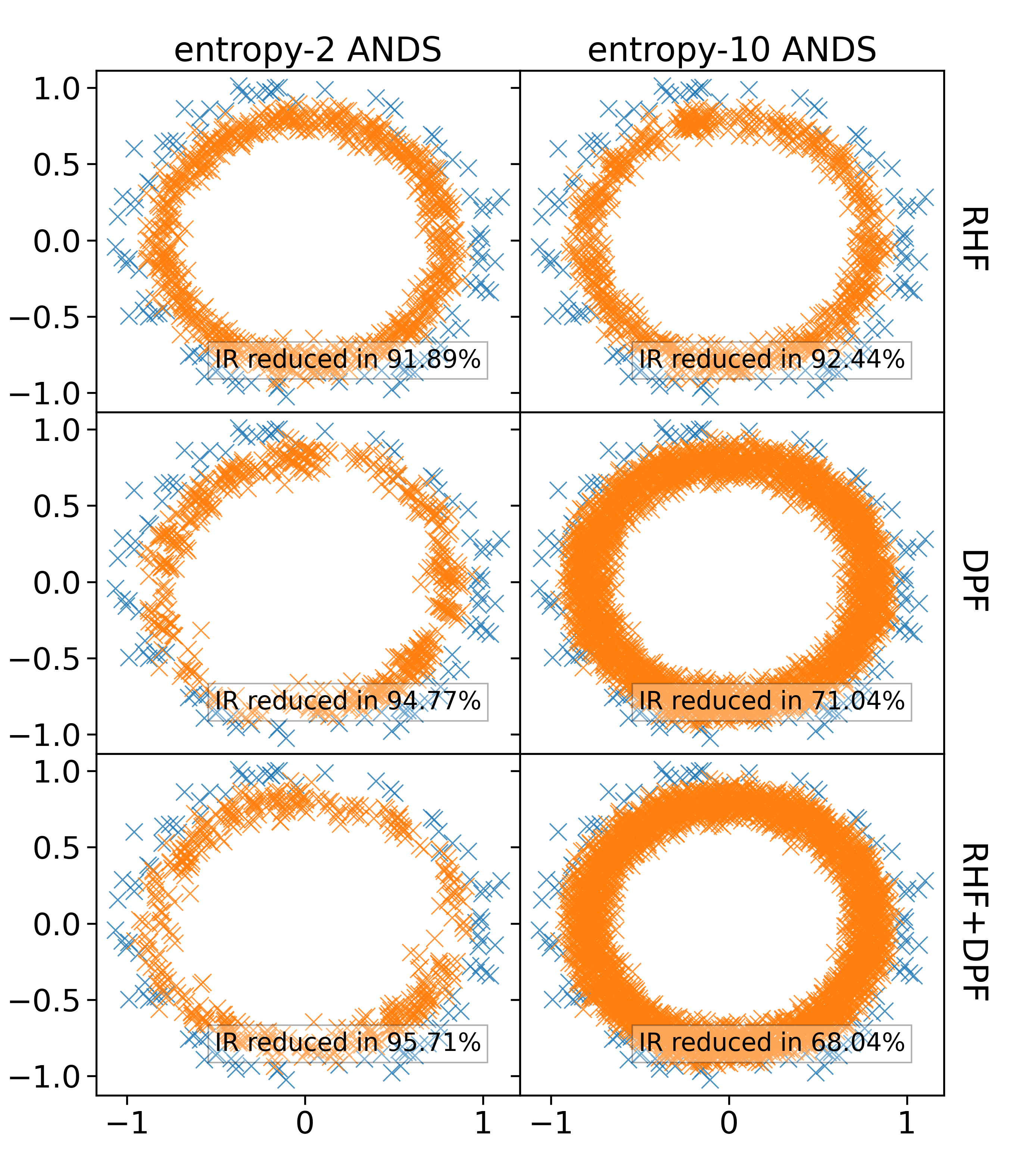}
		\caption{Effect of Entropy-based sampling in an inner-circles dataset with $IR=100$}
		\label{Fig:sampling_example1b}
	\end{subfigure}
\hfill
	\begin{subfigure}[b]{0.23\textwidth}
		\centering
		\includegraphics[scale=0.15]{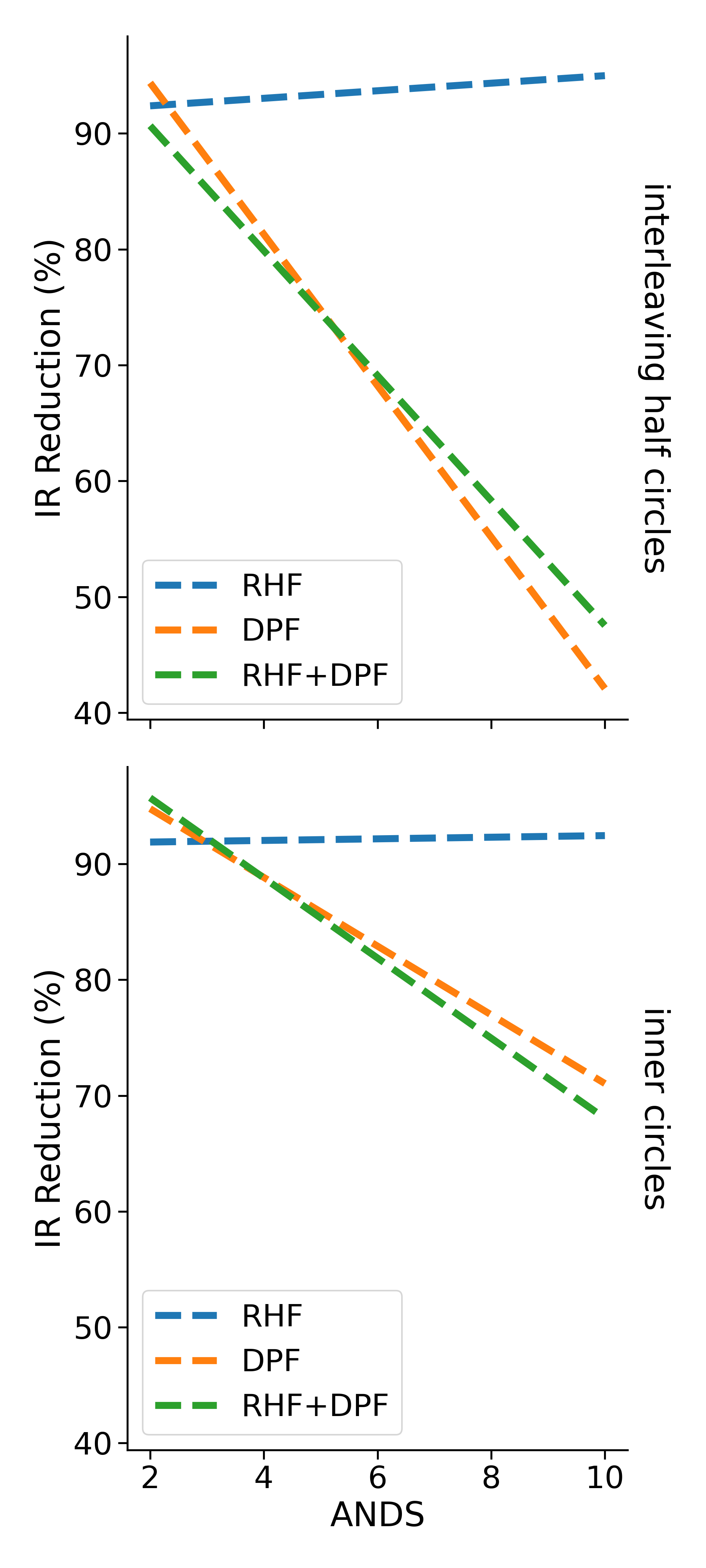}
		\caption{Effect of ANDS in the IR using Entropy-based sampling}
		\label{Fig:sampling_example1c}
	\end{subfigure}
\caption{Visualizing effects of Entropy-based sampling algorithm in 2 toy datasets with $IR=100$}
\label{Fig:sampling_example1}
\end{figure*}

To understand the effects of the Entropy-based sampling in the instances selected, Figure \ref{Fig:sampling_example1} shows the difference of the instances preserved depending on the number of ANDs functions and the LSH families used. These preliminary insights demonstrate the influence and flexibility offered by the LSH parameters to achieve specific behaviors in the sampling.  The two datasets are created simulating an $IR=100$. Interestingly, in Figure \ref{Fig:sampling_example1c}, where the $\%$ of IR reduction is shown along with the LSH families (line color) and the number of ANDs, the inverse relationship between ANDs and the IR reduction is only exhibited in the $DPF$ and $RHF+DPF$ families. In contrast, in the $RHF$ family the number of ANDs does not have a relevant influence; for the inner circles dataset, the IR reduction is almost constant, and for the interleaving half-circles dataset, a slightly direct relationship is observed. Figures \ref{Fig:sampling_example1a} and \ref{Fig:sampling_example1b} visually confirm the different effects of the number of ANDs and families in the IR of the datasets.

\subsection{DROP3 Sampling Method}\label{section:drop3}

Notwithstanding entropy provides a fast alternative for IS with minimum knowledge of the sample's distribution, clearly, it is not using information about the structure of the data. As mentioned before, DROP3 is one of the most well-known methods to perform IS, and it has already been used in a divide-and-conquer fashion. The present work follows the same idea of using DROP3 within the buckets created using LSH; however, the original DROP3 does not contemplate edges cases presented when the algorithm is applied to a specific subset of high imbalanced samples. Therefore, in this work, we propose three modifications of the standard DROP3 algorithm, which are included in the algorithm \ref{alg:drop3_unbalanced} and explained next. 


\begin{algorithm}[h!t!]
   \DontPrintSemicolon
	\SetKwData{Left}{left}\SetKwData{This}{this}\SetKwData{Up}{up}
	\SetKwFunction{Union}{Union}\SetKwFunction{FindCompress}{FindCompress}
	\SetKwInOut{Input}{input}\SetKwInOut{Output}{output}
    
    \Input{$X_{u}$ = instances in bucket $u$, $mode$ = how to deal with one class bucket,  $k = \#$ of neighbors}
    
	\Output{$S_{u}$ = selected instances $S_{u} \subset X_{u}$}
	
	$S_{u} \longleftarrow X_{u}$\;
	
	\uIf{labels of $S_{u}$ are majority class $\ \land \ mode=one$ } 

	  selects randomly one sample $s$ and $S_{u} = s$  
	  \label{alg:drop3_unbalanced:mode1}
	
   \uElseIf{labels of $S_{u}$ are majority class $\ \land \ modes=boundaries$ }{
   
       $\mu \longleftarrow$ centroid of $S$\;
       $|D| \longleftarrow$ absolute distances $d \hspace{0.1cm} \forall s \in S_{u}$ to $\mu$\;
       $\sigma_{D} \longleftarrow$ standard deviation of $|D|$
    	  
    	  \uIf{$\sigma_{D} > 0.001 $ }{
    	 
    	    \ForEach{instance  $s \in S_{u}$}{
    	  
    	        \eIf{ $\mu + \sigma_{D} \geq d  \geq \mu - \sigma_{D}$}{pass}
    	        {Remove $s$ from $S_{u}$ }\label{alg:drop3_unbalanced:mode2}
	    }
	 }
   }
	
   \uElse{
       Sort instances in $S_{u}$ by distances to their nearest enemy\;
        \ForEach{instance $s \in S_{u}$}{
        	Find the k+1 nearest neighbors of $s$ in $S_{u}$\;
            Add $s$ \text{its neighbors list of associates}\;
        }
        \ForEach{instance $s \in S_{u}$}{
           \label{alg:drop3_unbalanced:nodeletes}
        	\uIf{$(s \in majority \ class) $}{
    			$with \longleftarrow$ $\#$ associates of $s$ classified correctly with $s$ as a neighbor\;
                $without \longleftarrow$ $\#$ associates of $s$ classified correctly without $s$\;
                \uIf{$without \geq with$}{
                    \ForEach{associate $a \in S_{u}$}{
                        Remove $s$ from $a$'s list of neighbors\; 
                        Find new neighbor $n_{a}$ for $a$\;
                        Add $a$ to $n_{a}$ associate list\; 
                    }
                Remove $s$ from $S_{u}$\;
                }
          	}
         } 
     }
     \Return $S_{u}$
 \caption{Modified DROP3-sampling}
  \label{alg:drop3_unbalanced}
\end{algorithm}

\subsubsection{DROP3 modifications}

\begin{figure*}[!h!t]

	\begin{subfigure}[b]{0.45\textwidth}
		\centering
		\includegraphics[scale=0.12]{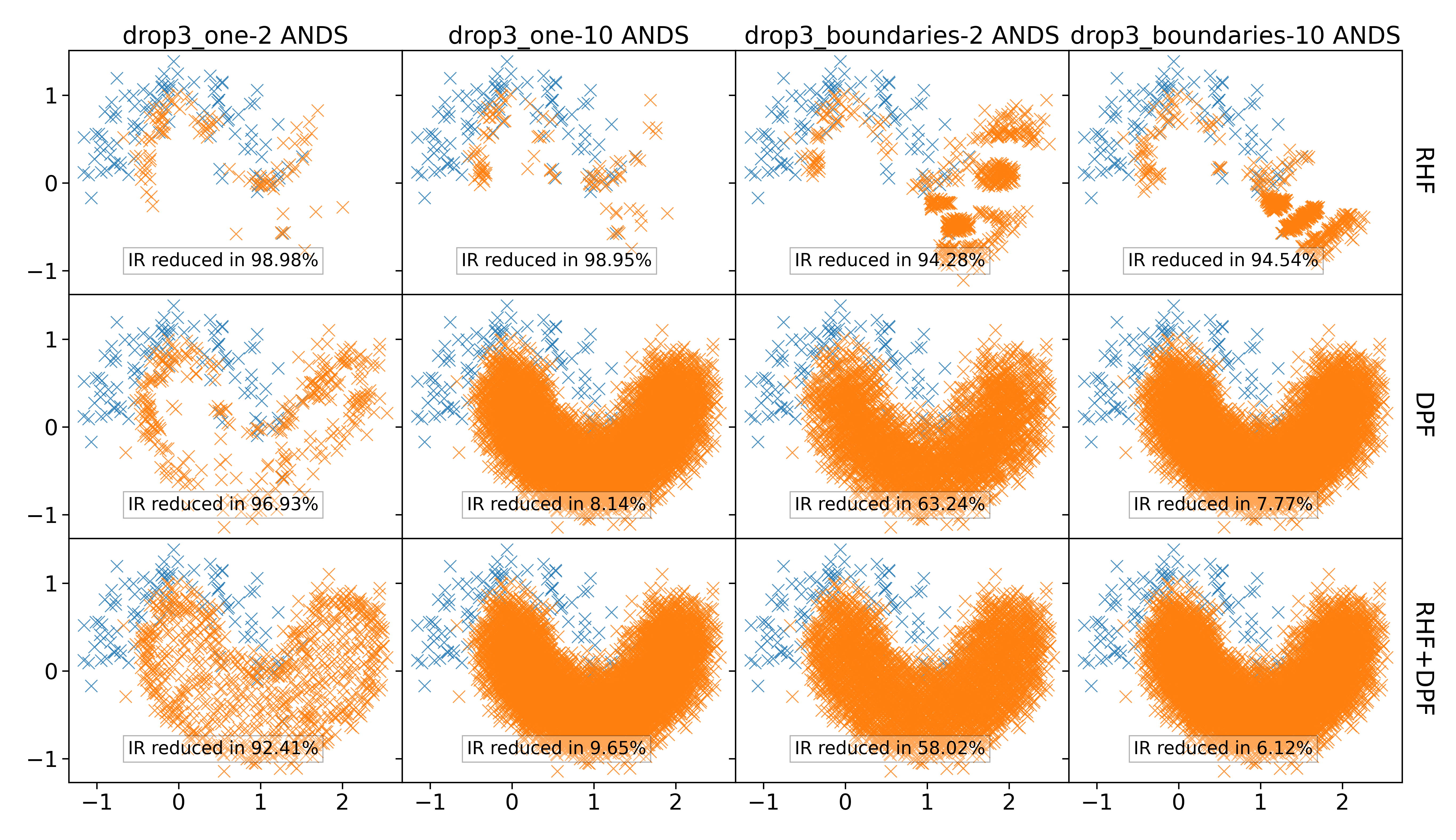}
		\caption{Effect of modified DROP3-sampling in an interleaving half-circles dataset with $IR=100$}
		\label{Fig:sampling_example2a}
	\end{subfigure}
\hfill
	\begin{subfigure}[b]{0.45\textwidth}
		\centering
		\includegraphics[scale=0.12]{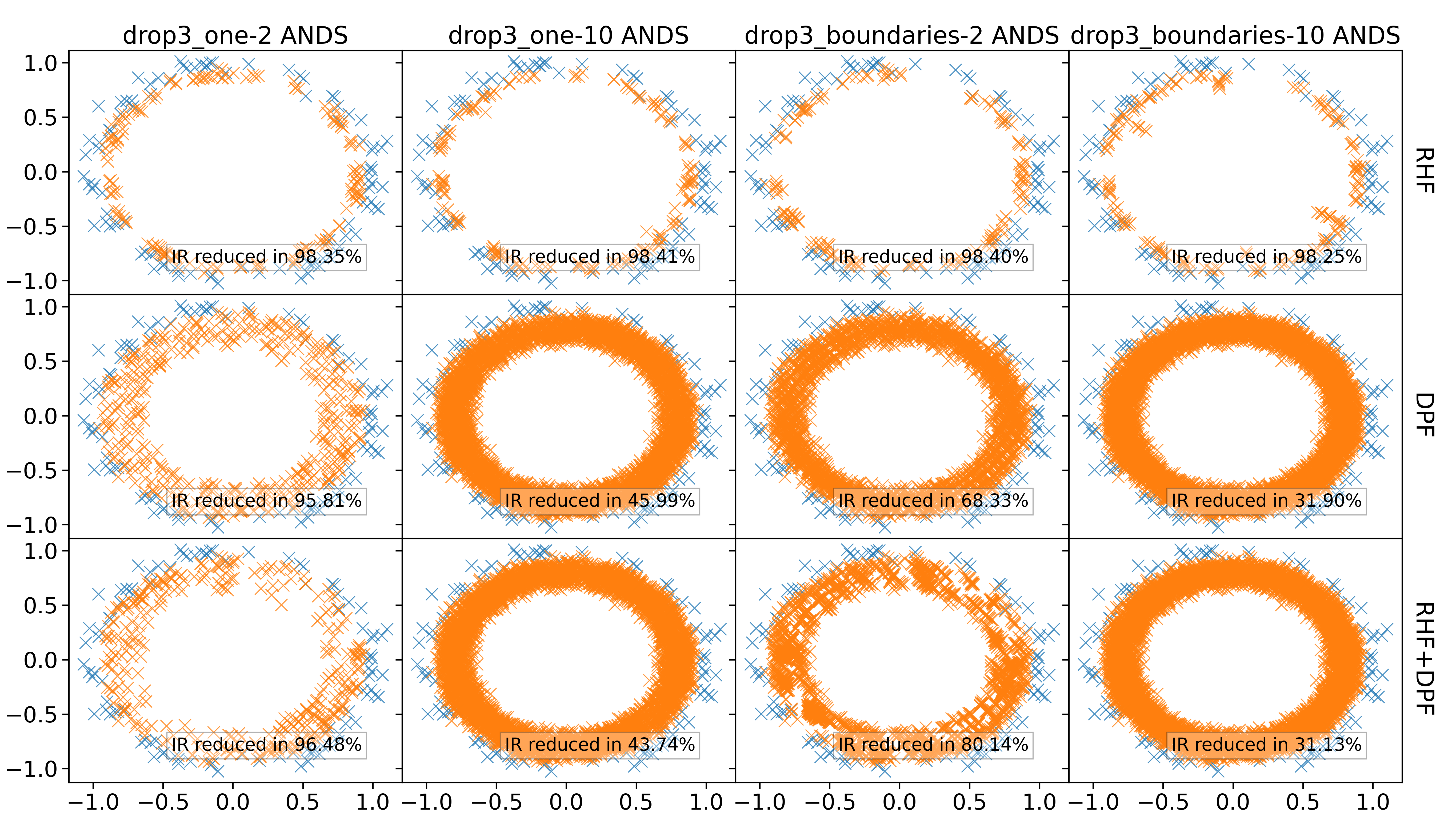}
		\caption{Effect of modified DROP3-sampling in an inner-circles dataset with $IR=100$}
		\label{Fig:sampling_example2b}
	\end{subfigure}
\centering	
\begin{subfigure}[b]{0.7\textwidth}
		\centering
		\includegraphics[scale=0.14]{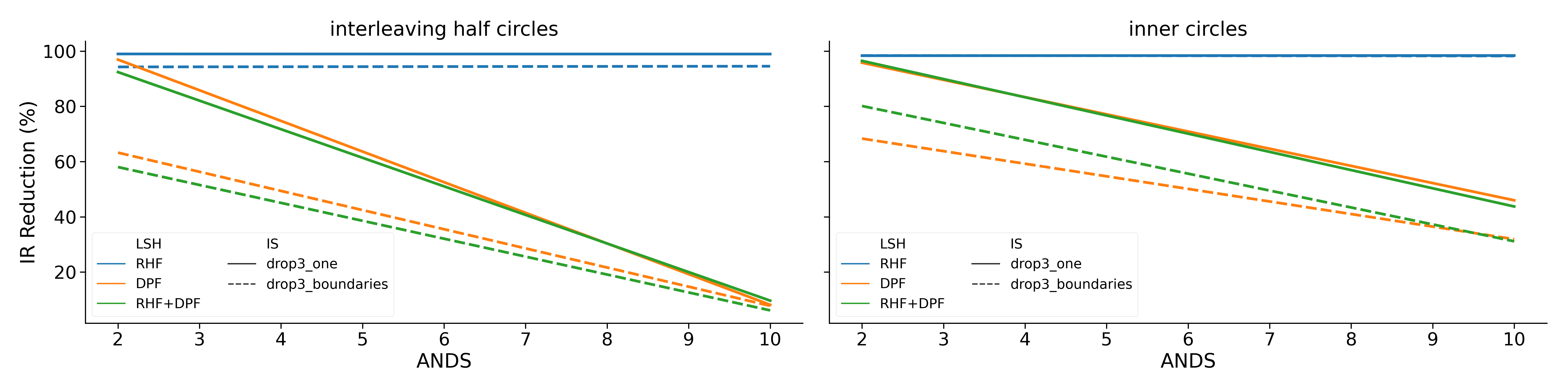}
		\caption{Effect of ANDS in the IR using modified DROP3-sampling}
		\label{Fig:sampling_example2c}
	\end{subfigure}
\caption{Visualizing effects of modified DROP3-sampling algorithm in 2 toy-datasets with $IR=100$}
\label{Fig:sampling_example2}
\end{figure*}

\begin{itemize}
    \item Originally, DROP3 proposed a noise-filtering step, where instances misclassified by a $k$-nearest neighbour($k$-NN) model are removed. However, the implementation proposed in this work does not apply such a noise-filtering step due to the high cost of training a $k$-NN in every single bucket. Moreover, its relevance is diminished since in a imbalanced context the minority class is unrepresented. Applying the $k$-NN filter could lead to eliminate minority class samples before actually start the instance selection process. 
    \item The second modification ensures the DROP3 never deletes a minority class instance due to the high cost of this action in imbalanced contexts. 
    \item The last modification introduces a $mode$ parameter that determines how to treat a bucket with samples of only one class. When  $mode = ``one"$, the algorithm randomly selects one unique sample to represent the entire bucket; instead, when  $mode =  ``boundaries"$, the algorithm preserves the instances located at the boundaries of the bucket. The process is performed by computing the bucket's mean ($\mu$) and removing the samples inside the radius $\mu \pm \sigma$, where $\sigma$ is the standard deviation of the distance of all samples to the centroid $\mu$. However, if $\sigma<0.001$, only the centroid is kept to avoid unnecessary calculations.
    
\end{itemize}  
Using the same toy-datasets of Section \ref{section:entropy}, Figure \ref{Fig:sampling_example2} shows the difference of the instances persisted depending on the number of ANDs functions, the LSH families used, and the value of $mode$ parameter for the modified DROP3-sampling. Similarly to the behavior observed in Figure \ref{Fig:sampling_example1}, the $RHF$ family performs differently than the other two families; along with the number of ANDs tested and both $mode$ values, the IR reduction is constant. Notice that the IR reduction is less for the interleaving half-circles dataset when $mode = boundaries$. 
The inverse relationship between ANDs and the IR reduction is also presented in the $DPF$ and $RHF+DPF$ families, but using $mode=one$ can achieve more significant reductions on the IR; However,  these IR reduction differences between the $mode$ values are notably more minor as the number of ANDs increases.



\subsection{Base Model}\label{section:base_model}
For the sake of comparison, the evaluation of the ML approaches whether a dataset was pre-processed using the proposed IS methods or not, was assessed using a Random Forest (RF) \cite{LeoBreiman2001} as the base model. RF was selected mainly because it provides fairly good results in non-linear problems and its training process allows parallel and distributed implementations (such as the one in Apache Spark), so it is suitable for Big Data Context. In the RF Apache Spark implementation, the model has two  hyperparameters: the number of trees ($B$) in the forest and the maximum depth of each tree. 


\section{Experiments and results}\label{section:experiments}

\begin{figure}[!h!b!t!]
\centering
	\includegraphics[scale=0.14]{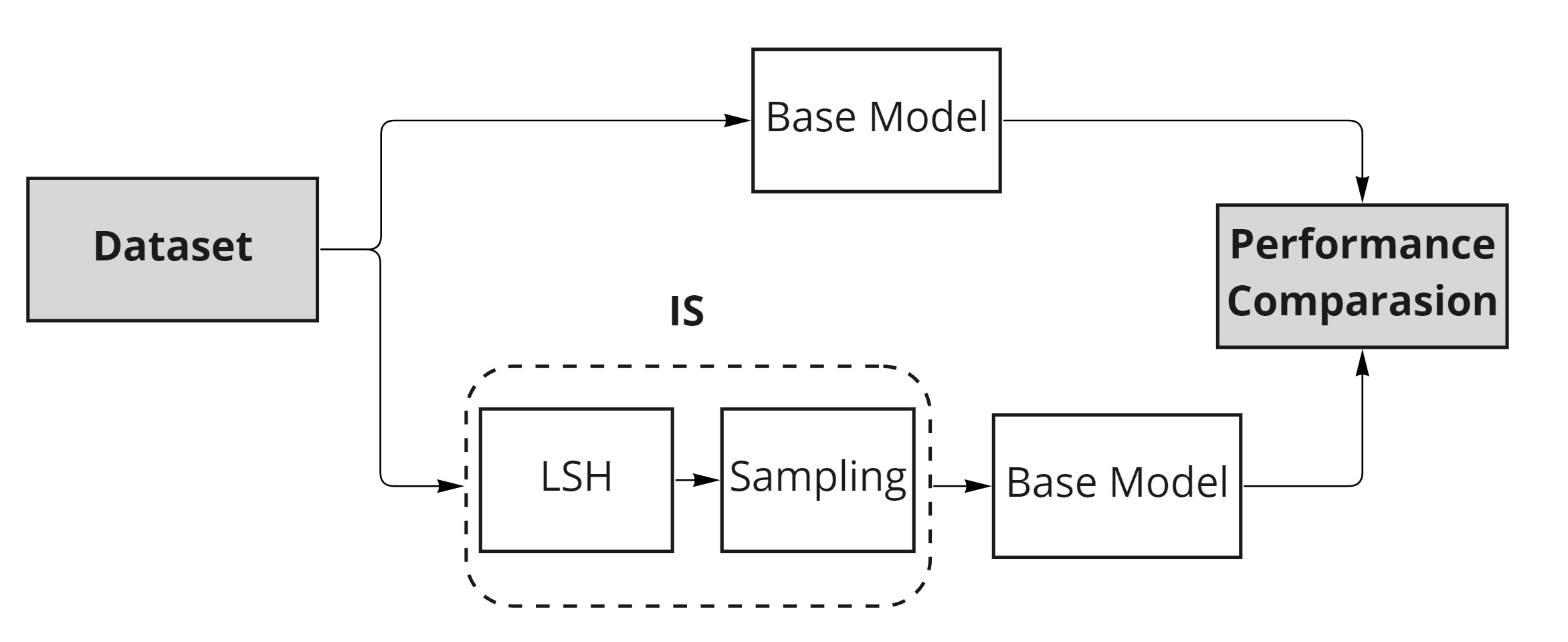}
	\caption{Methodology proposed to evaluate the effect of LSH as Instance Selection method}
	\label{Fig:Methodology}
\end{figure}

The figure \ref{Fig:Methodology} shows a summary of the methodology followed in this work to evaluate the effect of IS methods in different ML tasks. All the experiments were executed using two high IR datasets and one popular imbalanced dataset. For each dataset, a baseline model was trained without using IS and their performance metrics compared with those obtained using the different variants of IS methods described in section \ref{section:lsh}.

For the sake of performing comparisons fairly, experiments did not include the original DROP3 algorithm since it is sequential by nature with a computational cost of $O(n^{3})$ \cite{jankowski2004comparison}, which makes it unfeasible for the size of some of the dataset used for testing in this work. This complexity results from the iterative way it searches the $k$-neighborhoods to decide if the sample examined has to be removed. Moreover, a distributed implementation of any member of the DROP algorithm family will require modifications of the original approach to take advantage of the current frameworks' distributed nature, but this would change its essence.

\subsection{Datasets}\label{subsection:datasets}

Table \ref{tab:dataset} describes the three datasets used. To the best of our knowledge, the first dataset is the only one with high IR, and a large volume of samples publicly available  \cite{DalPozzolo:2014, Carcillo:2019}; it consists of samples from anonymized credit card transactions\cite{kaggle}. The second one is the Page Blocks dataset \cite{Dua:2019}, which has been used as a benchmark in several works in the context of IS  methods \cite{fernandez2017, Derrac2012, Garcia-Osorio2010}; although the dataset does not satisfy the restrictions about data size and IR, this dataset is a good reference for comparison purposes. The last dataset used was granted to an investigation under a non-disclosure agreement. The data set consists of $133$ features derived from credit card transactions from a payment gateway and contains more than 6 million samples.

\begin{table}[!h!b!t]
\centering

\caption{Datasets descriptions. The Column ``Samples + class'' is the number of samples in the minority class.}
\label{tab:dataset}

\begin{tabular}{ccccc@{}}
\toprule
Database & IR & \begin{tabular}[c]{@{}l@{}}Samples \\ + class\end{tabular} & \begin{tabular}[c]{@{}l@{}}Total \\  Samples\end{tabular} & \ Features \\ \midrule
Creditcard      & 577 & 492  & 285299 & 29 \\
pageblocks     & 9 & 559 & 5473 &  10\\
gateway &  1376 & 4715 & 6493035 &133  \\ \bottomrule
\end{tabular}
\end{table}

\subsection{Experimental setup}\label{subsection:experimentalsetup}

The training and validation procedures explained in Figure \ref{Fig:Methodology} are carried out using a five-folds stratified cross-validation strategy. Only training subsets are subject to sample selection for the experiments where any of the proposed IS methods is used; test sets are kept unchanged. Considering the class imbalance of the datasets, the metrics selected to compare the performance of the different IS variants, and the baseline model, are mainly the Balanced Accuracy ($BAcc$) and the $Gmean$ between the sensibility($Se$) and the specificity($Sp$). $Se$ and $Sp$ are metrics that evaluate the effectiveness of the classifier in each of the two classes separately; $BAcc$ and $Gmean$ are better alternatives in this case due to the well-known problem of using the Accuracy in high IR problems \cite{Kuncheva2019}. $BAcc$ and $Gmean$ are defined as:

\begin{equation}\label{eq:gmean}
Gmean =  \sqrt{Se\cdot Sp}
\end{equation}
\begin{equation}\label{eq:bacc}
BAcc  =  \frac{Se+Sp}{2}
\end{equation}
where $Se$ and $Sp$ are given by:
\begin{equation}\label{eq:sensitivity}
Se=\frac{TP}{TP+TN} \quad Sp=\frac{TP}{TP+TN}
\end{equation}
and $TP$ stands for True positives, $FP$ for False Positives, $TN$ for true negatives, and $FN$ for false negatives.
It was also included the F-measure ($F1$) that is helpful to compare the models when $Gmean$ is similar. $F1$ is defined as: 
\begin{equation}\label{eq:others}
F1 =  \frac{2\cdot TP}{2\cdot TP + FP + FN}
\end{equation}

The performance metrics and training-sample size reduction obtained after IS application were compared using a Friedman test \cite{Friedman1937} following the Iman–Davenport modification \cite{ImanDavenport1980} to assess if there are significant differences among the algorithm configurations. The Friedman test is the non-parametric alternative to the one-way ANOVA with repeated measures. It is used for continuous data that has violated the assumptions necessary to run the one-way ANOVA with repeated measures (e.g., data that has marked deviations from normality) and use the average ranks of the results of the IS configurations. After the test results analysis, the tradeoff between the reduction of the dataset and the performance was also studied.\\

Subsequently, the Krusal Wallis Test \cite{Kruskal1952} was used to determine if the IS results were significantly different from the base model. Then, a Wilcoxon signed-rank test \cite{Wilcoxon1945} was used to assess the significance of the differences found on each configuration of IS algorithms with the base model. In these multiple pairwise tests, the Wilcoxon p-value was adjusted using the Holm procedure to counteract the problem of multiple comparisons \cite{Holm1979}. Based on the results of this last test, only the IS methods that showed significant differences from the base model were selected to determine which IS configuration had better performance. These final comparisons were performed using the Wilcoxon signed-rank test.\\ 


In all the cases, the RF hyperparameters were adjusted according to the following grids: $B$ taking values in the interval $[10,25,50]$ and depths in $[10,20,30]$. The type of LSH family (RHF, DPF, and RHF+DPF), the IS method (entropy, drop-one, and drop-boundaries), and the number of ANDs were also adjusted during experiments.  The grid of possible values for the number of ANDs was set to $[2,4,6,8,10]$. The number of neighbors $k$ in the DROP3-based sampling methods was set to three. All the experiments were performed in the Apache Spark Framework\cite{spark}.\\

Furthermore, to assess the computational cost of the implementations and their capacity to handle large volumes of data, horizontal and vertical scalability experiments were carried out. The first one consists of the variation of the number of executors of Apache Spark and measuring the time spent to perform the IS, leaving the dataset's size constant; the numbers of executors tested were $[6,8,10,12,14,16]$. The vertical scalability experiments consisted of the dataset size variation, leaving a fixed number of executors ($12$). The data-set size was varied according to the following proportions: $[0.2,0.4,0.6,0.8,1.0]$. For the sake of comparison with further works, the scalability experiments were performed using the Creditcard dataset.

Finally, the experiments were run in a cluster with 3 Nodes with the following hardware specification: one node with 12 Intel(R) Xeon(R) CPU E5-2603 v3 with 370 GB in memory ram, other with 20 Intel(R) Xeon(R) CPU E5-2603 v3 with 309 GB in memory ram, and the third node with 20 Intel(R) Xeon(R) CPU E5-2620 v2 with 185 GB in memory ram. The software, frameworks, and hardware was managed by Cloudera Manager\cite{cdh}.

\subsection{Results}\label{subsec:results}

\begin{figure*}[!h!t]
\centering
	\includegraphics[scale=0.19]{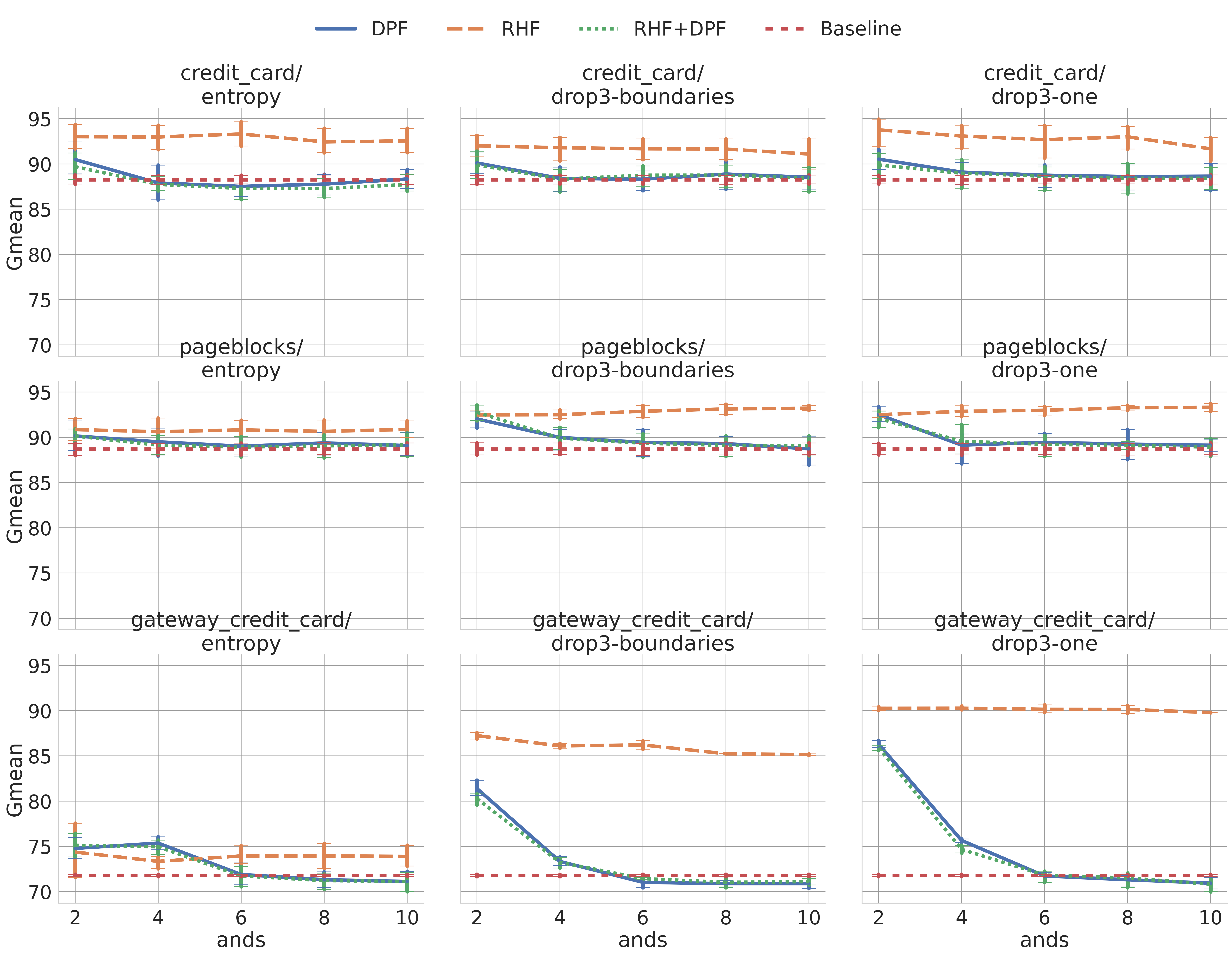}
	\caption{Performance of the proposed IS methods in terms of prediction effect.}
\label{Fig:Gmean}
\end{figure*}

Figure \ref{Fig:Gmean} shows the $Gmean$ prediction measure for each IS method applied to the three datasets described in section \ref{subsection:datasets} and compared to the baseline model.  The figure also shows the performance for the different LSH families evaluated and the number of ANDs functions. Most of the results show IS algorithms overcome the baseline model in all of the datasets, even though the success depends on the IS parameters, including the LSH family and the sampling method. For example, the entropy sampling method exhibits few differences in the gateway dataset, and the number of ANDs or the LSH family does not significantly affect the performance. However, in the DROP3 sampling methods, the number of ANDs can undermine the benefits of IS, specially for the $DPF$ and $RHF+DPF$ families. Oppositely, the $RHF$ family is not impacted in the same way when the number of ANDs changes. In the other datasets, the LSH family display similar patterns but the differences are notably less than for the gateway dataset. It is also interesting to notice that the effectiveness of the IS algorithm with $RHF$ family and the three sampling methods increases according to the IR; in datasets with higher IR, the differences of the performance metric compared with the base model are evident.\\

\begin{figure*}[!h!t]
\centering
	\includegraphics[scale=0.2]{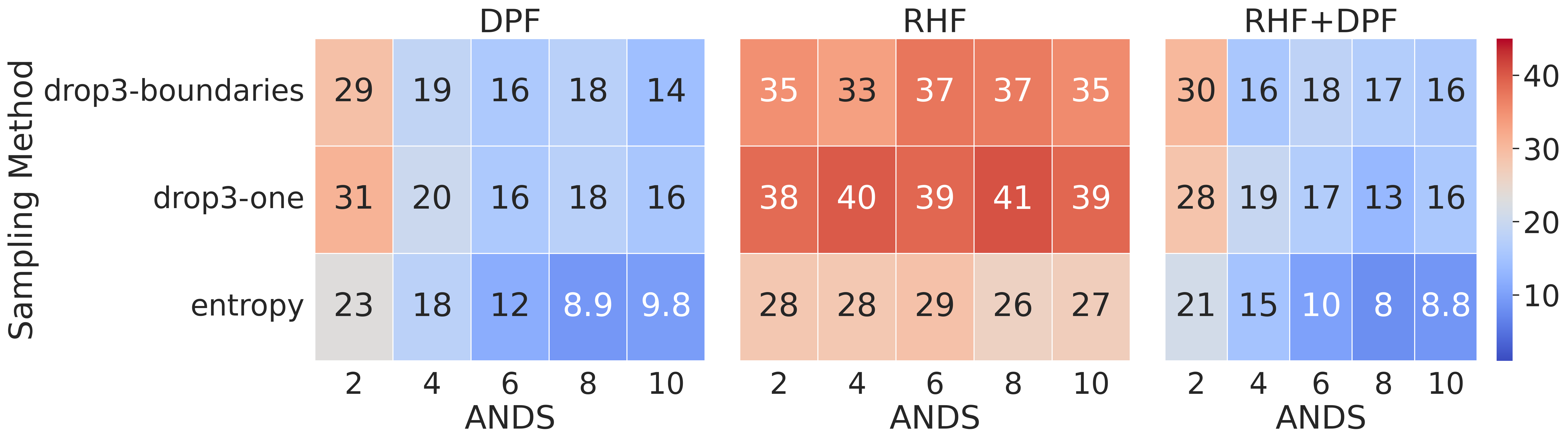}
	\caption{Results Ranks for Friedman test with p-value: 0.0000 and F-value = 14.6197. The redder the box, the higher the rank achieved by the IS/LSH family method.}
	\label{Fig:ranktest}
\end{figure*}

Although Figure \ref{Fig:Gmean} helps visually to understand the different influences of the IS in the performance, the statistical test helps to assess the significance of those results. For instance,  the results of the Friedman test yielded an F-Value of $14.6197$ with a p-value = $0.0000$, which means that all the configurations tested belong to different distributions; thus, the IS configuration chosen have crucial effects in the final results. Figure \ref{Fig:ranktest} shows the ranks of the configurations. According to the number of configurations tested, the worst possible rank is $1$, and the best is $45$. Therefore, the rank of the Friedman test can be seen as a metric to assess which method delivered better performance. In this preliminary evaluation, there is evidence that the $RHF$ family outperforms the others, the ranks of that family are higher than $26$ and, in some cases, closer to the maximum rank possible. In contrast, in other families, in most of the configurations, the ranks are less than $20$; the only exception is when $2$ ANDs were used. This latter observation shows more evidence of the significant impact of ANDs on the algorithm's performance.


\begin{figure*}[!h!t]
\centering
	\includegraphics[scale=0.06]{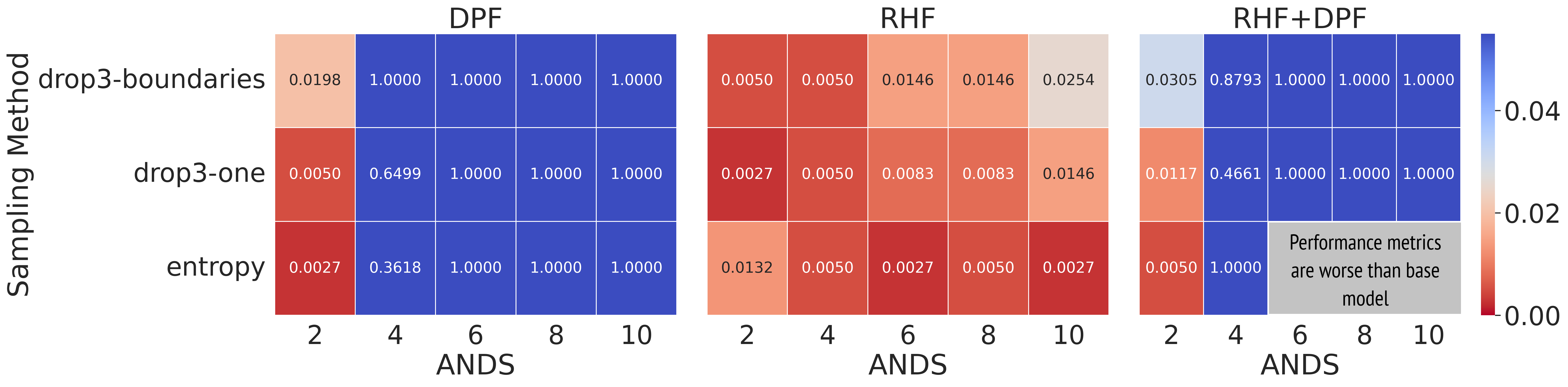}
	\caption{P-values of a pairwise comparison of proposed methods with the baseline model. Red cells means the method was significant better than the reference.}
	\label{Fig:referece_comp}
\end{figure*}

Knowing that the IS algorithms' configurations are significantly different, the  Kruskal-Wallis test was applied to estimate if IS performance results are significantly different from the performances obtained by the base model. The results of the test comparisons with the base model after the application of a Holm process are shown in Figure \ref{Fig:referece_comp}. The plot only displays the configurations that achieve a better performance than the base model; the p-value determines whether the difference seen is significant or not. The analysis of the results with $0.05$ as a critical p-value supports previous findings that the $RHF$ family outperforms the base model in all the number of ANDs tested. For the other two families, the configuration with $2$ ANDs is the only one that significantly differentiates against the baseline model. Accordingly, only the results of the $RHF$ were selected to determine if there are significant performance variations among the sampling methods.

In general, the data distributions in classification problems is expected to form clouds of samples or clusters inhabited mainly by samples of one class. After a typical standardization of variables, the clusters of different classes are expected to be located in opposite positions in relation to the origin of the feature space, so the results obtained by the $RHF$ family can be explained by the differences in how these families divide the feature space (see section \ref{section:lsh-functions}). In other words, the $DPF$ and $DPF+RHF$ families create buckets more prone to split the feature space into groups where the classes are mixed; on the other hand,  the $RHF$ tends to create regions where it is more probable to have samples of one class.


\begin{figure}[!h!b]
\centering
	\includegraphics[scale=0.3]{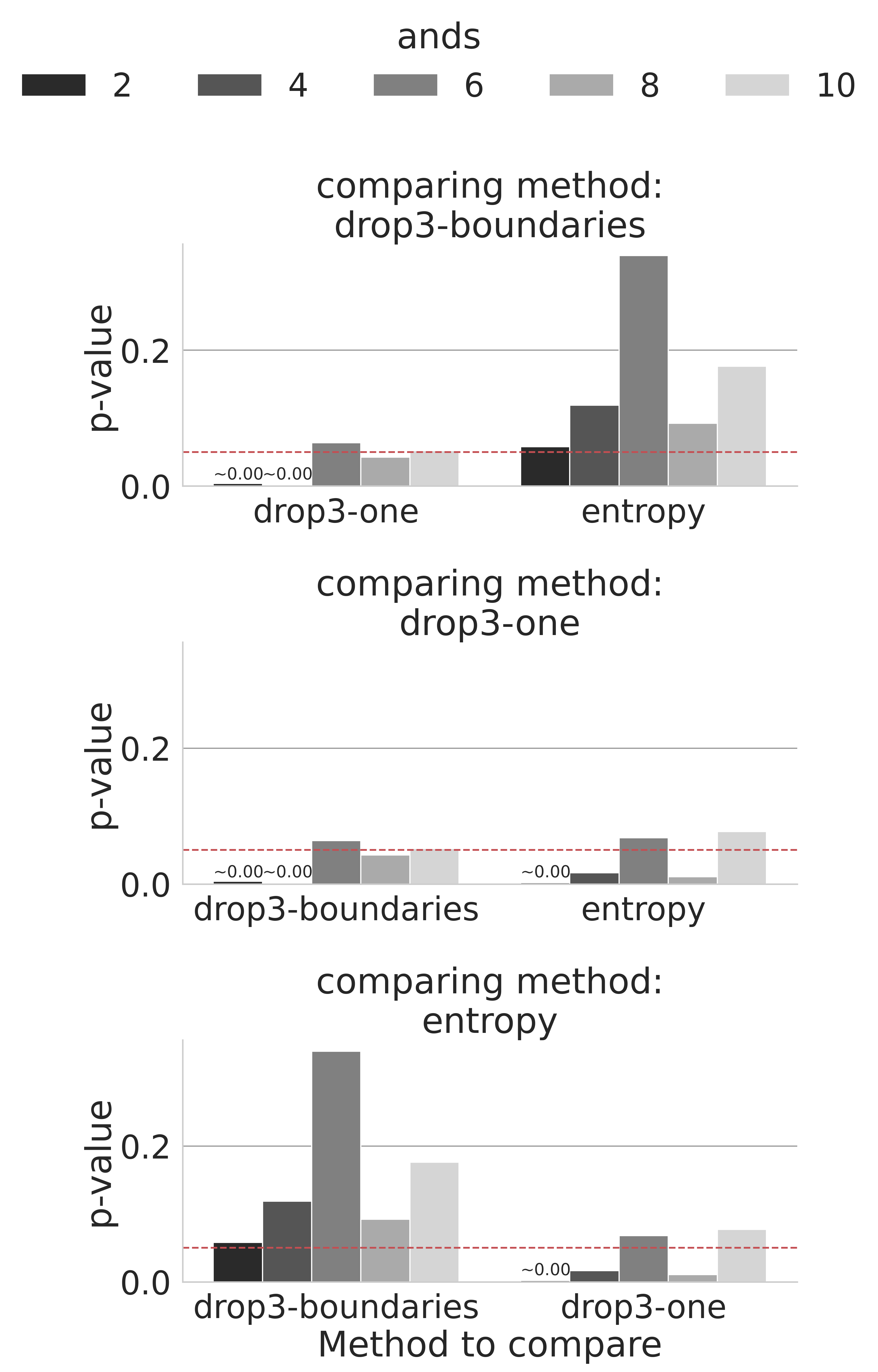}
	\caption{P-values of the pairswise wilcoxon comparasion tests of sampling methods with $RHF$ LSH family. The red line indicates critical p-value of 0.05}
	\label{Fig:final_comp}
\end{figure}

A final Wilcoxon test is applied to compare and determine if the differences observed among the sampling methods are significant. In this case, no Holm procedure to adjust the p-values is needed because the comparison is only for the pairs; thus, the results of  Wilcoxon tests are shown in Figure \ref{Fig:final_comp}. The test results proved that the drop3-one sampling method was significantly better in three of the five values of AND configurations; however, the correct number of ANDs cannot be determined intuitively. For example, the differences between the sampling methods configured with $6$ and $10$ ANDS were not significant. Ergo, selecting the number of ANDs is vital to deliver correct results, and it has to be tuned according to the problem.\\


\begin{figure}[!h!t]
\centering
	\includegraphics[scale=0.14]{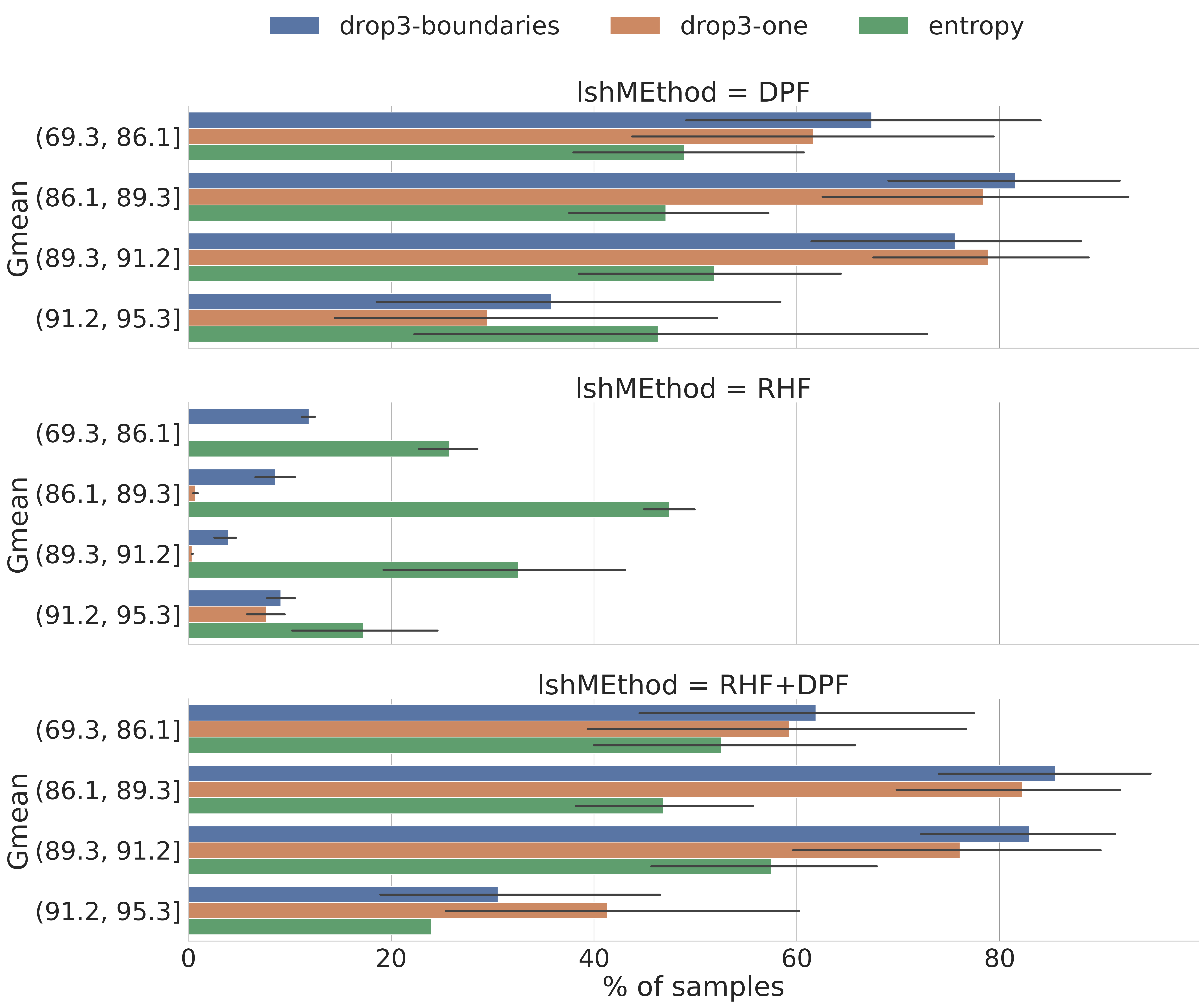}
	\caption{Relation between the $Gmean$ and the $\%$ of samples selected. Friedman test with  $F=114.0123$ and $p=0.0000$}
	\label{Fig:gmean_vs_reduction}
\end{figure}

The previous discussions and results have shown the effects of IS in terms of performance. Nevertheless, the IS methods aim to reduce the size of the original data sets without harming the performance, so for that reason, in Figure \ref{Fig:gmean_vs_reduction} the relationship between the $\%$ of samples used in training and the $Gmean$ achieved in testing is presented. The results generally show that the $Gmean$ is better when the IS methods selected fewer samples on the training subset. The $RHF$ family shows remarkable $Gmean$ values, keeping less than 20 $\%$ of samples in most cases. On the other hand, the $DPF$  and $RHF+DPF$ families produced similar $Gmean$ values selecting more than $40\%$ of training samples. In all the cases, the selection of the LSH family influences the tradeoff between performance and data size; for example, the drop3 sampling methods got higher $Gmean$ and fewer samples only in $RHF$, in other families, the Entropy-sampling method was better balancing such tradeoff. The Friedman test got a $F=114.0123$ with $p=0.0000$, indicating all the configurations of LSH and sampling methods showed significant differences in reducing the original dataset.\\


To summarize the findings and compare the final results of the best IS configuration, Table \ref{tab:final_comp} shows the two sets of comparison: best configuration by each dataset according to the $Gmean$, and best configuration according to the $\%$ of samples used in training. In the same table, the complementary performance metrics and the best baseline are presented. As we have analyzed before, the drop3-one with the $RHF$ family is the IS configuration with the best results. Furthermore, the requirement to tune the ANDs to fit the specific necessities of the problem is confirmed; the best value of this parameter varies depending on the dataset. The complementary metrics revealed that the IS methods outperform the baseline model in most cases except for $Sp$ in all datasets and the $F1$ only in the gateway credit card dataset. Still, the IS algorithms achieved values very close to the base model.

\begin{table*}[!h!t]\centering
\captionsetup{justification=centering}
\caption{Final results of best parameters of LSH+IS for each dataset compared with the best baseline. C1 = RHF+DROP-ONE with 2 ANDS. C2 = RHF+DROP3-ONE with 10 ANDS. C3 = RHF+DROP3-ONE with 4 ANDS. The column names with *  corresponds to the baseline metrics}\label{tab:final_comp}
\centering
\small
\resizebox{0.97\textwidth}{!}{
\begin{tabular}{p{1.2cm}p{1.5cm}p{1.7cm}p{0.95cm}p{0.95cm}p{0.95cm}p{0.95cm}p{0.95cm}p{1.5cm}p{0.95cm}p{0.95cm}p{0.95cm}p{0.95cm}p{0.95cm}}\toprule
\textbf{database} &\textbf{Parameters} & $\%$ \textbf{samples} &\textbf{Gmean} &\textbf{Gmean*} &\textbf{Se} &\textbf{Se*} &\textbf{Sp} &\textbf{Sp*} &\textbf{F1} &\textbf{F1*} &\textbf{BAcc} &\textbf{BAcc*} & \\\midrule
credit card &C1 &0.47 $\pm$ 0.01 &\textbf{93.77} $\pm$ 2.06 &88.26  $\pm$  1.27 &\textbf{88.09} $\pm$ 3.82 &78.02 $\pm$ 2.25 &99.85 $\pm$ 0.01 &\textbf{99.87} $\pm$ 0.01 &\textbf{63.77} $\pm$ 1.03 &60.93 $\pm$ 0.68 &\textbf{93.97} $\pm$ 1.91 &88.94 $\pm$ 1.13 \\ \hline
pageblocks &C2 &15.46 $\pm$ 0.26 &\textbf{93.32} $\pm$ 0.56 &88.71 $\pm$ 1.69 &\textbf{98.02} $\pm$ 0.77 &86.61 $\pm$ 3.78 &88.85 $\pm$ 1.17 &\textbf{90.91} $\pm$ 0.74 &\textbf{66.22} $\pm$ 0.18 &63.37 $\pm$ 1.01 &\textbf{93.44} $\pm$ 0.54 &88.76 $\pm$ 1.64 \\ \hline
gateway credit card &C3 &0.30$\pm$0.00 &\textbf{90.27} $\pm$ 0.23 &71.79 $\pm$ 0.3 &\textbf{81.54} $\pm$ 0.42 &51.53 $\pm$ 0.43 &99.94 $\pm$ 0.00 &\textbf{100.00} $\pm$ 0.00 &61.99 $\pm$ 0.12 &\textbf{67.26} $\pm$ 0.31 &\textbf{90.74} $\pm$ 0.21 &75.77 $\pm$ 0.21  \\
\midrule
\multicolumn{14}{c}{\textbf{best results using the \% of samples and performance} }\\
\midrule
credit card &C3 &0.47 $\pm$ 0.01 &93.09 $\pm$ 1.58 &88.26 $\pm$ 1.27 &86.80 $\pm$ 2.93 &78.02 $\pm$ 2.25 &99.85 $\pm$ 0.01 &99.87 $\pm$ 0.01 &63.44 $\pm$ 0.79 &60.93 $\pm$ 0.68 &93.33 $\pm$ 1.46 &88.94 $\pm$ 1.13 \\ \hline
pageblocks &C1 &11.81 $\pm$ 0.33 &92.5 $\pm$ 0.49 &88.71 $\pm$ 1.69 &97.63 $\pm$ 1.12 &86.61 $\pm$ 3.78 &87.65 $\pm$ 1.37 &90.91 $\pm$ 0.74 &66.13 $\pm$ 0.26 &63.37 $\pm$ 1.01 &92.64 $\pm$ 0.47 &88.76 $\pm$ 1.64 \\ \hline
gateway credit card &C1 &0.30 $\pm$ 0.00 &90.27 $\pm$ 0.26 &71.79 $\pm$ 0.30 &81.53 $\pm$ 0.46 &51.53 $\pm$ 0.43 &99.94 $\pm$ 0.00 &100.0 $\pm$ 0.00 &61.99 $\pm$ 0.13 &67.26 $\pm$ 0.31 &90.73 $\pm$ 0.23 &75.77 $\pm$ 0.21 \\
\bottomrule
\end{tabular}}
\end{table*}

\begin{figure}[!h!t]
\centering
	\includegraphics[scale=0.19]{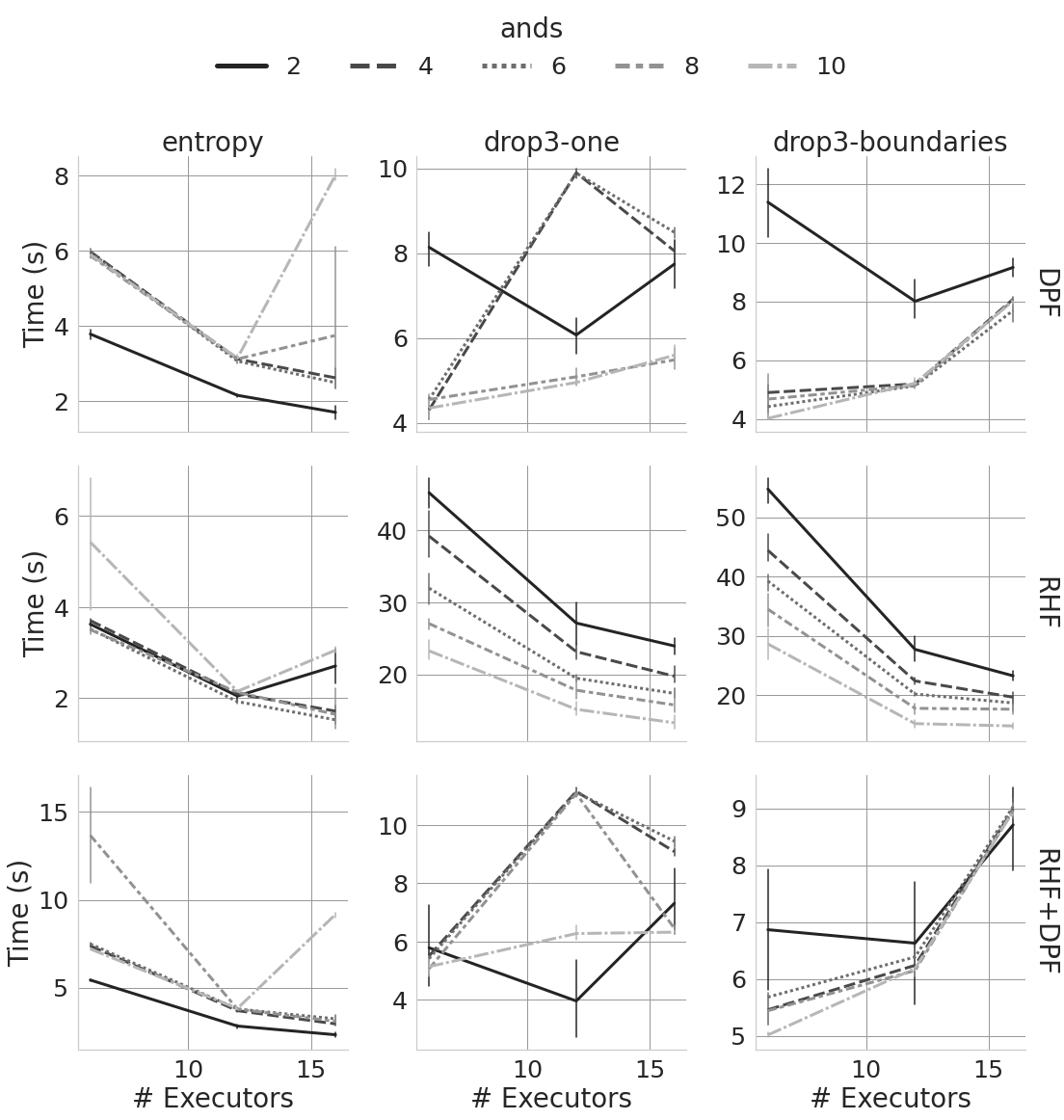}
	\caption{Horizontal scalability results}
	\label{Fig:horizontal}
\end{figure}

\begin{figure}[!h!t]
\centering
	\includegraphics[scale=0.17]{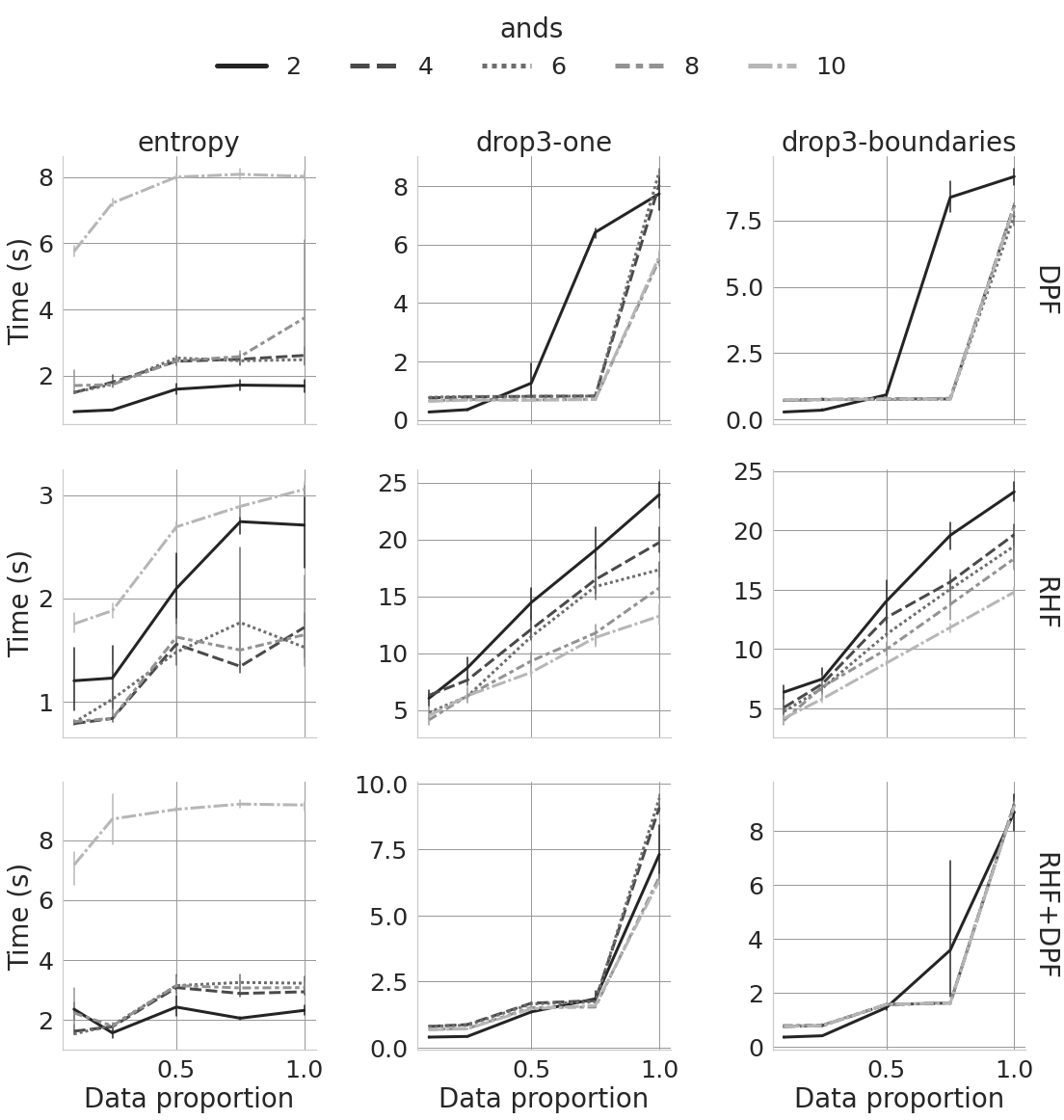}
	\caption{vertical scalability results}
	\label{Fig:vertical}
\end{figure}

Finally, after testing and demonstrating the effects of algorithms in the performance on high IR problems, it is needed to analyze the scalability potential of the algorithms proposed. Figure \ref{Fig:horizontal}  and Figure \ref{Fig:vertical} shows respectively, the results of the horizontal and vertical scalability experiments. \\

In the ideal behavior of horizontal scalability, the execution time should decrease linearly according to the number of executors. The results showed that the $RHF$ with the three sampling methods exhibited an ideal pattern in most cases. However, it is impacted by the number of ANDs; in the Entropy sampling case with $2$ and $10$ ANDs, the execution time increases after $12$ executors.  In other families, the patterns are more inconsistent. For instance, in the $DPF$ family using entropy-sampling, the execution time displays a linear tendency in all the ANDs configurations until $12$ executors. However, after that number, the configuration with $8$ and $10$ ANDs increases execution time. The tendency changes could be a consequence of the structure dataset used; the sample density could be unevenly forcing the current algorithm implementation to make irregular partitions of the data and increasing the execution time.\\


Similarly, the drop3-boundaries preserve their linear tendency until $12$ executors and after that number, the execution time increases for all the ANDs configurations tested. On another hand, the drop3-one has a more heterogeneous behavior, for $2$, $4$, and $6$ ANDs, the execution time has both direct and inverse relationships in a different $\#$ of executors, but for $8$ and $10$ ANDs displays contrary patterns from the ideal. For the $RHF+DPF$ family, the patterns observed are very similar to the $DPF$ family. However, in drop3-one, any ANDs configuration manifests a direct relationship with the execution time. Depending on executors' $\#$, that relationship changed from direct to inverse and vice versa. In general terms, it is interesting to notice that the entropy sampling method using the families $DPF$ and $RHF$ displays less computational costs without considering the number of ANDs. When the family was $DPF+RHF$, the entropy-based method displays results comparable with DROP3 using boundaries, but it is heavily dependant on the configuration: number of ANDs and executors.\\

Generally speaking, if an algorithm shows robust vertical scalability, it is expected that the algorithm's execution time increases linearly according to the size of data processed. The results revealed that the proposed IS methods manifested the ideal pattern when $RHF$ is used.  On the contrary, the differences between the two DROP3 methods are not visible, and the $DPF$ and $RHF+DPF$ families did not display robust behaviors. For example, in entropy sampling, the execution time is more influenced by the number of ANDs than the data size increments. On the other hand, the drop3 time execution behaves positively and directly to the data size processed; however, the slope of results changes depending on the data proportion, suggesting a non-linear relation. Finally, in this vertical scalability experiment and similarly to the horizontal scalability test, entropy sampling also presents the fewest computational costs among the methods proposed.


\section{Conclusions}\label{section:conclusions}

The work proposed three novel IS methods adapted to handle high IR problems in big data contexts. The algorithms were developed in the Apache Spark framework, which guarantees their scalability. The results showed that the three IS methods achieved significant performance improvements compared to a base model using the whole dataset. In terms of $Gmean$, the best IS method achieved improvements between $5\%$ and $19\%$ in absolute terms, with a slight loss in $Sp$ and $F1$. These improvements are dependent on the parameters of the algorithms. The results demonstrated that the $RHF$ outperforms the other two LSH families studied, achieving, in general, higher $Gmean$ values using $20\%$ of the training subset. However, the results also showed that the number of ANDs is a parameter that needs to be tunned according to the problem at hand.\\

Similarly, the results demonstrated that the modifications proposed to DROP3 achieve outstanding results, especially the drop3-one sampling method exhibited the best results when combined with the $RHF$ family. However, depending on the context, entropy sampling could be a good option if the execution time has a higher cost in the particular problem. The experiments executed to prove the robustness of algorithms' scalability and implementation have shown the sensibility of the methods to their parameters. Depending on the LSH families, the number of ANDs, and the sampling method used, the vertical or the horizontal scalability can show a more or less robust behavior.\\

Future work is suggested to deeply study the relation of vertical and horizontal scalability to the samples density distribution. Furthermore, these experiments could originate other modifications of the algorithms to decouple their robustness to its parameters, such as detecting the regions with a higher density of samples and applying the techniques recursively to maintain the algorithm's robustness. In the same way, the authors observed the necessity to compare the algorithms proposed with other state-of-the-art methods such as OligoIS, LSH-IS-S, and LSH-IS-F; however, implementations efforts are needed to adapt those methods to big data frameworks.

\section*{Acknowledgments}

This work was partially supported by Ruta N and the University of Antioquia under the grant 020C-2016. J. Arias-Londoño started this work at the Antioquia University and finished it supported by a María Zambrano grant from the Universidad Politécnica de Madrid, Spain.
\bibliography{article}

\begin{thebibliography}{10}
\expandafter\ifx\csname url\endcsname\relax
  \def\url#1{\texttt{#1}}\fi
\expandafter\ifx\csname urlprefix\endcsname\relax\def\urlprefix{URL }\fi
\expandafter\ifx\csname href\endcsname\relax
  \def\href#1#2{#2} \def\path#1{#1}\fi

\bibitem{Halevy2009}
A.~Halevy, P.~Norvig, F.~Pereira, {The unreasonable effectiveness of data},
  IEEE Intelligent Systems 24~(2) (2009) 8--12.

\bibitem{bishop:2006:PRML}
C.~M. Bishop, Pattern Recognition and Machine Learning, Springer, 2006.

\bibitem{rasmussen2006gaussian}
C.~E. Rasmussen, C.~K. Williams, Gaussian process for machine learning, MIT
  press, 2006.

\bibitem{arnaiz2016instance}
{\'A}.~Arnaiz-Gonz{\'a}lez, J.-F. D{\'\i}ez-Pastor, J.~J. Rodr{\'\i}guez,
  C.~Garc{\'\i}a-Osorio, Instance selection of linear complexity for big data,
  Knowledge-Based Systems 107 (2016) 83--95.

\bibitem{Chang2011}
E.~Y. Chang, K.~Zhu, H.~Wang, H.~Bai, {PSVM : Parallelizing Support Vector
  Machines on Distributed Computers}, In Foundations of Large-Scale Multimedia
  Information Management and Retrieval~(2) (2011) 213--230.

\bibitem{math8020286}
H.~Saadatfar, S.~Khosravi, J.~H. Joloudari, A.~Mosavi, S.~Shamshirband, A new
  k-nearest neighbors classifier for big data based on efficient data pruning,
  Mathematics 8~(2).

\bibitem{Snelson2005}
E.~Snelson, Z.~Ghahramani, {Sparse Gaussian Processes using}, Advances in
  Neural Information Processing Systems 18 (NIPS 2005) (2005) 1--8.

\bibitem{Garcia-Osorio2010}
C.~Garc{\'{i}}a-Osorio, A.~de~Haro-Garc{\'{i}}a, N.~Garc{\'{i}}a-Pedrajas,
  {Democratic instance selection: A linear complexity instance selection
  algorithm based on classifier ensemble concepts}, Artificial Intelligence
  174~(5-6) (2010) 410--441.

\bibitem{JoelLusCarboneraB2017}
{Joel Luıs Carbonera(B)}, {An Efficient Approach for Instance Selection}, in:
  L.~Bellatreche, S.~Chakravarthy (Eds.), Big Data Analytics and Knowledge
  Discovery, Vol.~1 of Lecture Notes in Computer Science, Springer
  International Publishing, Cham, 2017, pp. 228----243.

\bibitem{Garcia-Pedrajas2013}
N.~Garc{\'{i}}a-Pedrajas, J.~Peŕez-Rodr{\'{i}}guez, A.~{De Haro-Garci{\'{a}}},
  {OligoIS: Scalable instance selection for class-imbalanced data sets}, IEEE
  Transactions on Cybernetics 43~(1) (2013) 332--346.

\bibitem{Krawczyk2016}
B.~Krawczyk, {Learning from imbalanced data: open challenges and future
  directions}, Progress in Artificial Intelligence 5~(4) (2016) 221--232.

\bibitem{Leevy:2018}
J.~L. Leevy, T.~M. Khoshgoftaar, R.~A. Bauder, N.~Seliya, {A survey on
  addressing high-class imbalance in big data}, Journal of Big Data 5~(1).

\bibitem{Fernandez2019}
A.~Fern{\'{a}}ndez, S.~Garc{\'{i}}a, M.~Galar, R.~C. Prati, {Learning from
  Imbalanced Data Sets (2018, Springer International Publishing).pdf}, 2019.

\bibitem{Kuncheva2019}
L.~I. Kuncheva, {\'{A}}.~Arnaiz-Gonz{\'{a}}lez, J.~F. D{\'{i}}ez-Pastor, I.~A.
  Gunn, {Instance selection improves geometric mean accuracy: a study on
  imbalanced data classification}, Progress in Artificial Intelligence 8~(2)
  (2019) 215--228.

\bibitem{Garcia-Pedrajas2014}
N.~Garc{\'{i}}a-Pedrajas, A.~{De Haro-Garc{\'{i}}a}, {Boosting instance
  selection algorithms}, Knowledge-Based Systems 67 (2014) 342--360.
\newblock \href {http://dx.doi.org/10.1016/j.knosys.2014.04.021}
  {\path{doi:10.1016/j.knosys.2014.04.021}}.

\bibitem{Wilson2000}
D.~R. Wilson, T.~R. Martinez, Reduction techniques for instance-based learning
  algorithms, Machine Learning 38~(3) (2000) 257--286.

\bibitem{Brighton2002}
H.~Brighton, C.~Mellish, {Advances in instance selection for instance-based
  learning algorithms}, Data Mining and Knowledge Discovery 6~(2) (2002)
  153--172.

\bibitem{Arnaiz-Gonzalez2017}
{\'{A}}.~Arnaiz-Gonz{\'{a}}lez, A.~Gonz{\'{a}}lez-Rogel, J.~F.
  D{\'{i}}ez-Pastor, C.~L{\'{o}}pez-Nozal, {MR-DIS: democratic instance
  selection for big data by MapReduce}, Progress in Artificial Intelligence
  6~(3) (2017) 211--219.

\bibitem{Czarnowski2018}
I.~Czarnowski, P.~J{\k{e}}drzejowicz, Cluster-based instance selection for the
  imbalanced data classification, in: International Conference on Computational
  Collective Intelligence, Springer, 2018, pp. 191--200.

\bibitem{Tsai2019}
C.-F. Tsai, W.-C. Lin, Y.-H. Hu, G.-T. Yao, Under-sampling class imbalanced
  datasets by combining clustering analysis and instance selection, Information
  Sciences 477 (2019) 47--54.

\bibitem{DavidW.Aha1911}
{David W. Aha}, {Dennis Kibler}, M.~K. Albert, {Instance-Based Learning
  Algorithms DAVID}, Machine learning 6 (1911) 37----66.

\bibitem{spark}
{Apache Software Foundation}, \href{https://spark.apache.org}{Apache spark
  2.4.0} (2018).
\newline\urlprefix\url{https://spark.apache.org}

\bibitem{Indyk1998}
P.~Indyk, R.~Motwani, {Approximate nearest neighbors}, in: Proceedings of the
  Thirtieth Annual ACM Symposium on Theory of Computing, Association for
  Computing Machinery, 1998, pp. 604--613.

\bibitem{rajaraman2012mining}
A.~Rajaraman, J.~D. Ullman, Mining of massive datasets, Cambridge University
  Press, Cambridge, 2012.

\bibitem{datar2004locality}
M.~Datar, N.~Immorlica, P.~Indyk, V.~S. Mirrokni, Locality-sensitive hashing
  scheme based on p-stable distributions, in: Proceedings of the twentieth
  annual symposium on Computational geometry, ACM, 2004, pp. 253--262.

\bibitem{ivanchykhin2017regular}
D.~Ivanchykhin, S.~Ignatchenko, D.~Lemire, Regular and almost universal
  hashing: an efficient implementation, Software: Practice and Experience
  47~(10) (2017) 1299--1323.

\bibitem{LeoBreiman2001}
{Leo }, {Random Forests}, Machine Learning 45~(1) (2001) 5--32.

\bibitem{jankowski2004comparison}
N.~Jankowski, M.~Grochowski, Comparison of instances seletion algorithms i.
  algorithms survey, in: International conference on artificial intelligence
  and soft computing, Springer, 2004, pp. 598--603.

\bibitem{DalPozzolo:2014}
A.~{Dal Pozzolo}, O.~Caelen, Y.-A. {Le Borgne}, S.~Waterschoot, G.~Bontempi,
  Learned lessons in credit card fraud detection from a practitioner
  perspective, Expert Systems with Applications 41~(10) (2014) 4915--4928.

\bibitem{Carcillo:2019}
F.~Carcillo, Y.-A. Le~Borgne, O.~Caelen, Y.~Kessaci, F.~Oble, G.~Bontempi,
  Combining unsupervised and supervised learning in credit card fraud
  detection, Information Sciences.

\bibitem{kaggle}
{Kaggle Inc}, \href{https://www.kaggle.com/mlg-ulb/creditcardfraud}{Credit card
  fraud detection} (2021).
\newline\urlprefix\url{https://www.kaggle.com/mlg-ulb/creditcardfraud}

\bibitem{Dua:2019}
D.~Dua, C.~Graff, \href{http://archive.ics.uci.edu/ml}{{UCI} machine learning
  repository} (2017).
\newline\urlprefix\url{http://archive.ics.uci.edu/ml}

\bibitem{fernandez2017}
A.~Fernández, C.~J. Carmona, M.~José~del Jesus, F.~Herrera, A pareto-based
  ensemble with feature and instance selection for learning from multi-class
  imbalanced datasets, International Journal of Neural Systems 27~(06) (2017)
  1750028, pMID: 28633551.

\bibitem{Derrac2012}
J.~Derrac, C.~Cornelis, S.~Garc{\'{i}}a, F.~Herrera, {Enhancing evolutionary
  instance selection algorithms by means of fuzzy rough set based feature
  selection}, Information Sciences 186~(1) (2012) 73--92.

\bibitem{Friedman1937}
M.~Friedman, The use of ranks to avoid the assumption of normality implicit in
  the analysis of variance, Journal of the American Statistical Association
  32~(200) (1937) 675--701.

\bibitem{ImanDavenport1980}
R.~L. Iman, J.~M. Davenport, Approximations of the critical region of the
  fbietkan statistic, Communications in Statistics - Theory and Methods 9~(6)
  (1980) 571--595.
\newblock \href
  {http://arxiv.org/abs/https://doi.org/10.1080/03610928008827904}
  {\path{arXiv:https://doi.org/10.1080/03610928008827904}}.

\bibitem{Kruskal1952}
W.~H. Kruskal, W.~A. Wallis, Use of ranks in one-criterion variance analysis,
  Journal of the American Statistical Association 47~(260) (1952) 583--621.

\bibitem{Wilcoxon1945}
F.~Wilcoxon, Individual comparisons by ranking methods, Biometrics Bulletin
  1~(6) (1945) 80--83.

\bibitem{Holm1979}
S.~Holm, A simple sequentially rejective multiple test procedure, Scandinavian
  Journal of Statistics 6~(2) (1979) 65--70.

\bibitem{cdh}
{Clodudera Manager}, Cloudera manager 6.3.0 (2017).

\end{thebibliography}

\end{document}